\documentclass[journal]{IEEEtran}
\usepackage{booktabs}       % professional-quality tables
\usepackage{amsfonts}       % blackboard math symbols
\usepackage{graphicx}
\usepackage{subcaption}
\ifCLASSINFOpdf
\else
\fi
\usepackage{amsmath}
\usepackage{array}

\usepackage{stfloats}
\usepackage{url}
\newcommand{\vx}{{\mathbf{x}}}
\newcommand{\vk}{{\mathbf{k}}}
\newcommand{\vy}{{\mathbf{y}}}

\newcommand{\vz}{{\mathbf{z}}}

\newcommand{\vs}{{\mathbf{s}}}
\newcommand{\bbR}{{\mathbb{R}}}

\newcommand{\gp}{{\mathcal{GP}}}
\newcommand{\N}{{\mathcal{N}}}
\newcommand{\F}{{\mathcal{F}}}

\newcommand{\freq}[1]{\hat{#1}} % frequency domain
\newcommand{\timedelay}{{\theta}}
\newcommand{\phase}{{\phi}}
\newcommand{\Finv}{{{\F}_{s\rightarrow \tau}^{-1}}}

\newcommand{\varn}{{{\sigma}^2_{n}}}
\newcommand{\Var}{{\Sigma}}

\newcommand{\tra}{^{\top}}
\newcommand{\vmu}{{\boldsymbol{\mu}}}
\newcommand{\vtime}{{\boldsymbol{\timedelay}}}
\newcommand{\vphase}{{\boldsymbol{\phase}}}
\newcommand{\imagi}{{\dot{\iota}}}
\newcommand{\uptext}{\overbrace}
\newcommand{\upline}{\overline}
\newcommand{\gcsm}{\text{GCSM}}
\newcommand{\mosm}{\text{MOSM}}
\newcommand{\mocsm}{\text{MOCSM}}
\newcommand{\sm}{\text{SM}}
\newcommand{\itimej}{{{i}\times{j}}}

\begin{document}
\title{Multi-Output Convolution Spectral Mixture for Gaussian Processes}

\author{Kai Chen,~\IEEEmembership{}
        Twan van Laarhoven,~\IEEEmembership{}
		Perry Groot,~\IEEEmembership{}
		Jinsong Chen,~\IEEEmembership{}
		and~Elena Marchiori~\IEEEmembership{}% <-this % stops a space

\thanks{Kai Chen is with the Shenzhen Institutes of Advanced Technology, Chinese Academy of Sciences, with the Shenzhen College of Advanced Technology, University of Chinese Academy of Sciences, Shenzhen 518055, China, and also with the Institute for Computing and Information Sciences, Radboud University, 6525 EC Nijmegen, The Netherlands (e-mail: kyoungchen@gmail.com).}% <-this % stops a space
\thanks{Twan van Laarhoven is with the Institute for Computing and Information Sciences, Radboud University, 6525 EC Nijmegen, and Faculty of Management, Science and Technology, Open University of The Netherlands, The Netherlands (e-mail: twan.vanlaarhoven@ou.nl).}%
\thanks{Perry Groot is with the Institute for Computing and Information Sciences, Radboud University, 6525 EC Nijmegen, The Netherlands (e-mail: perry.groot@science.ru.nl).}% <-this % stops a space
\thanks{Jinsong Chen (Co-corresponding author) is with the Shenzhen Institutes of Advanced Technology, Chinese Academy of Sciences, Shenzhen 518055, China (e-mail: js.chen@siat.ac.cn).}
\thanks{Elena Marchiori (Co-corresponding author) is with the Institute for Computing and Information Sciences, Radboud University, 6525 EC Nijmegen, The Netherlands (e-mail: elenam@cs.ru.nl).}}


\markboth{Journal of \LaTeX\ Class Files,~Vol.~14, No.~8, August~2015}%
{Shell \MakeLowercase{\textit{et al.}}: Bare Demo of IEEEtran.cls for IEEE Journals}



\maketitle

\begin{abstract}		
		Multi-output Gaussian processes (MOGPs) are an extension of Gaussian Processes (GPs) for predicting multiple output variables (also called channels, tasks) simultaneously. In this paper we use the convolution theorem to  design a new kernel for MOGPs, by modeling cross channel dependencies through cross convolution of time and phase delayed components in the spectral domain. The resulting kernel is called Multi-Output Convolution Spectral Mixture  (MOCSM) kernel. Results of extensive experiments on synthetic and real-life datasets demonstrate the advantages of the proposed kernel and its state of the art performance. MOCSM enjoys the desirable property to reduce to the well known Spectral Mixture (SM) kernel when a single-channel is considered. A comparison with the recently introduced Multi-Output Spectral Mixture kernel reveals that this is not the case for the latter kernel, which contains quadratic terms that generate undesirable scale effects  when the spectral densities of different channels are either very close or very far from each other in the frequency domain.  		
\end{abstract}


% \fi
\IEEEpeerreviewmaketitle

\section{Introduction}
\IEEEPARstart{G}{}{aussian} processes (GPs) \cite{Rasmussen2010,Rasmussen2006} are an elegant Bayesian approach to model an unknown function. They provide regression models where a posterior distribution over the unknown function is maintained as evidence is accumulated. This allows GPs to learn complex functions when a large amount of evidence is available, and  makes them robust against overfitting in the presence of little evidence. GPs can model a large class of phenomena through the choice of the kernel, which characterizes one's assumption on how the unknown function autocovaries \cite{Rasmussen2010}. The choice of the kernel is a core aspect of a GP design, since the posterior distribution can significantly vary for different kernels. As a consequence, various kernels, like Squared Exponential, Periodic, Mat\'ern, and kernel design methods, have been proposed (see, e.g., \cite{Rasmussen2006}).

The extension of GPs to predict multiple output variables (or channels, tasks) simultaneously is known as multi-output Gaussian processes (MOGPs). MOGPs are also called multi-task Gaussian processes (MTGPs). Therefore for convenience, in the sequel, MOGPs and MTGPs will be used inter-exchangeably. %alternatively. 
MOGPs not only model temporal or spatial relationships among infinitely many random variables, as in scalar GPs, but also account for the statistical dependence across different sources of data (called channels or tasks) \cite{Parra2017}. How to choose an appropriate kernel to jointly model the cross covariance between channels and auto-covariance within each channel is the core problem of MOGPs design.

Early approaches for this problem, like Linear Model of Coregionalization (LMC) \cite{Goovaerts1997} \cite{bonilla2008multi} \cite{Duerichen2015a}, consider linear combinations of shared kernels. More expressive methods like the multi-kernel method \cite{Melkumyan2011} and the convolved latent function framework \cite{Alvarez2009} consider convolution to construct cross covariance functions, and assume that each channel has its own kernel. Recently spectral mixture (SM) kernels have been used and extended for MOGPs, resulting in a principled methods for constructing cross covariance functions which are easier to interpret: the SM-LMC kernel \cite{Wilson2014,Wilson2011}, which uses shared components, and a more flexible kernel called cross Spectral Mixture (CSM) kernel \cite{Ulrich2015a}, which considers the power and phase correlations between multiple outputs. The CSM kernel, however, cannot capture time delayed cross correlations between channels. The more recent Multi-Output Spectral Mixture kernel (MOSM) \cite{Parra2017} addresses this limitation. 
However in the MOSM kernel, cross channel weights contain each channel weight coefficient as multiplicative term. This may induce amplified scale effects on cross weights when the spectral densities of different channels are either very close or very far from each other. Also,  MOSM does not reduce to the ordinary spectral mixture kernel (SM) when only a single channel is considered, since the channel weight term in MOSM occurs squared.

In order to address these drawbacks, we use the convolution theorem to  design a new kernel for MOGPs, by modeling cross channel dependencies through cross convolution of time and phase delayed components in the spectral domain. The resulting kernel is called Multi-Output Convolution Spectral Mixture  (MOCSM) kernel. Results of extensive experiments on synthetic and real-life datasets demonstrate the advantages of the proposed kernel and its state of the art performance. Also, MOCSM enjoys the desirable property to reduce to the well known Spectral Mixture (SM) kernel when a single-channel is considered.

The paper is structured as follows. In Section \ref{sec:fm} we show how to construct cross components with time and phase delay through convolution, we extend the cross components to the multi-output scenario, and compare MOCSM and MOSM. In Section \ref{sec:rw} we further analyze the differences between MOCSM and the aforementioned SM-based approaches for MOGPs.  Section \ref{sec:exp} describes experiments on synthetic and real dataset. Summary, concluding remarks and future work on this topic are given in the final Section \ref{sec:con}.

\section{Background}\label{bk}
We start with some background information on GPs, multi-output GPs, and spectral mixture kernels.
\subsection{Gaussian processes}
A Gaussian process defines a distribution over functions, specified by its mean  $m({\vx})$ and covariance  $k({\vx}, {{\vx}{'}})$ \cite{Rasmussen2006} for given input vector ${\vx}\in{\bbR}^{P}$. Thus we can define a GP as
\begin{align}
f({\vx})\sim{\gp}(m({\vx}), k({\vx}, {{\vx}{'}}))
\end{align}
Without loss of generality we assume the mean of a GP to be zero. The covariance function $k$ is applied to construct a positive definite covariance matrix on input points $X$, here denoted by $K=K({X}, {X})$. By placing a GP prior over functions through the choice of a kernel and parameter initialization, from the training data $X$ we can predict the unknown function value $\tilde{y}_*$ and its variance $\mathbb{V}[{\tilde{y}_*}]$ (that is, its uncertainty) for a test point ${\vx}_*$ using the following key predictive equations for GP regression \cite{Rasmussen2006}:
\begin{align}
\tilde{y}_*&=\vk_{*}\tra(K+{{\varn}}I)^{-1}{\vy}\\
\mathbb{V}[{\tilde{y}_*}]&=k({\vx}_*, {\vx}_*)-\vk_{*}\tra(K+{{\varn}}I)^{-1}\vk_{*}
\end{align}
where $\vk_{*}\tra$ is the covariances vector between ${\vx}_*$ and $X$, and ${\vy}$ are the observed values corresponding to $X$. Typically, the kernel function  contains free parameters ${\Theta}$, called hyper-parameters, which can be inferred by minimizing the Negative Log Marginal Likelihood (NLML) with respect to ${\Theta}$ and ${{\varn}}$, where the marginal likelihood $ p({\vy}|{\vx},{\Theta}) \sim {\N}(0, K_{\Theta}+{{\varn}}I)$ and ${{\varn}}$ is the noise level:
\begin{equation}\label{eq:nlml}
\begin{split}
\text{NLML}&=-\log\ p({\vy}|{\vx},{\Theta})\\
&\propto\uptext{\frac{1}{2}{\vy}\tra(K_{\Theta}+{{\varn}}I)^{-1}{\vy}}^{\text{model fit}}+ \uptext{\frac{1}{2}\log|K_{\Theta} + {{\varn}} I|}^{\text{complexity penalty}} + c,
\end{split}
\end{equation}
where $c$ is a normalization constant.

In multi-output GPs (MOGPs), we have multiple sources of data which specify related outputs. The construction of the MOGP covariance function $k_\text{MOGP}(\vx^i, \vx^j)$ models covariances of each channel and dependencies between pairs of different channels \cite{Genton2015} where $\vx^i$ and $\vx^j$ are respectively from $i$-th and $j$-th channel.

\subsection{Spectral mixture kernels}
Usually, the smoothness and generalization properties of GPs depend on the kernel function and its hyper-parameters. Therefore choosing a kernel and learning its hyper-parameters are the core steps of a GP.
Various kernels have been proposed \cite{Rasmussen2006}, such as Squared Exponential (SE), Periodic (PER), and general Mat\'ern (MA). Recently the Spectral Mixture (SM) kernels have been introduced \cite{Wilson2014,Wilson2013}, which are at the core of the extension we propose. We describe them briefly in the sequel. A stationary kernel is a function of $\tau=\vx - \vx'$, that is, it is invariant to translation of the input. From Bochner's Theorem (see e.g. \cite{Bochner2016}) a SM kernel, here denoted by  $k_\text{SM}$, is derived by modeling GP covariance functions via spectral densities that are scale-location mixtures of Gaussians. By considering the spectral density $\freq{k}$ to be a mixture of $Q$ Gaussians on ${\bbR}^{P}$ of the form ${{\freq{k}}_{{\sm}i}}({\vs})=[{\varphi}_{{\sm}i}({\vs}) + {\varphi}_{{\sm}i}(-{\vs})]/2$, where each ${\varphi}_{{\sm}i}({\vs})={\N}({\vs};\vmu_{i},{\Var}_{i})$ is a symmetrized scale-location Gaussian in the frequency domain, and by using inverse Fourier transform $\Finv$ of the spectral density, we have 
\begin{align}
\begin{split}
k_{\sm}(\tau)=&\Finv\bigg[\sum_{i=1}^Q{w_i}{{\freq{k}}_{{\sm}i}}({\vs})\bigg](\tau)\\
=&\sum_{i=1}^Q{w_i}\Finv\big[\big({\varphi}_{{\sm}i}({\vs}) + {\varphi}_{{\sm}i}(-{\vs})\big)/2\big](\tau).%\\
\end{split}
\end{align}
By applying the inverse Fourier transform we obtain
\begin{align}\label{eq:smp}
\begin{split}
k_{\sm}(\tau)&= \sum_{i=1}^Q{w_i}k_{{\sm}i}(\tau), \mbox{ where}
\end{split}\\
\begin{split}
k_{{\sm}i}(\tau)&={\cos\left(2\pi\tau\tra\vmu_{i}\right)}\prod_{p=1}^{P}{\exp\left(-2\pi^2\tau^{2}{\Var}_{i, p}\right)}.
\end{split}
\end{align}

Here $k_{{\sm}i}$ denotes the $i$-th Gaussian component, $w_i$, $\vmu_{i}=\left[\mu_{i,1},...,\mu_{i,P}\right]$, and ${\Var}_{i}=\text{diag}\left(\left[{\sigma_{i,1}^{2}},...,{\sigma_{i,P}^{2}}\right]\right)$ are weight, mean, and variance of the $i$-th component in frequency domain, respectively. The variance ${\sigma_{i}^{2}}$ can be interpreted as an inverse length-scale, $\mu_{i}$ as a frequency, and $w_i$ as the contribution (the signal variance %, the signal variance
) of the $i$-th component. 

\section{Multi-output convolution spectral mixtures}\label{sec:fm}
We now consider two natural extensions of spectral mixture kernels: (1) incorporation of time and phase delay dependencies between Gaussian components in a single-output setting; (2) incorporation of dependencies between channels in a multi-output setting. In the following section we describe our recent proposal for the first extension, which models time and phase delay dependencies between components through convolution.

\subsection{Modeling time and phase delay dependencies through convolution}
Recently the Generalized Convolution Spectral Mixture (GCSM) kernel was introduced \cite{2018arXiv180800560C,kai2019} to model time and phase delay dependencies (in a single task setting)  between Gaussian components $i$ and $j$. First, the SM kernel is enhanced by incorporating time and phase delay, which yields a generalized spectral density function. Next, the convolution theorem  is used (see e.g. \cite{kai2019,antoniou2016digital}), which states that multiplying the Fourier Transform (FT) of one function  with the complex conjugate of the other function gives the FT of their correlation. This theorem is used to decompose the (complex valued) spectral density into base spectral densities as follows: 
\begin{align}\label{eq:gcsm_spec}
{\freq{k}_{{\gcsm}}^{\itimej}({\vs})}=\hat{g}_{\text{GCSM}i}({\vs})\upline{\hat{g}_{\text{GCSM}j}}({\vs})
\end{align} where 
$$\hat{g}_{\text{GCSM}i}({\vs})=\sqrt{w_{i}\varphi_{\text{SM}i}({\vs})}\exp\left(-\pi{\dot{\iota}}(\boldsymbol{\theta}_{i}{{\vs}}+{\boldsymbol{\phi}_{i}})\right)
$$
is a base function of a complex-valued SM density with time delay $\vtime_{i}$ and phase delay $\vphase_{i}$. Here the overline %$\upline{-}$
denotes the complex conjugate operator. The resulting cross SM component $k_\text{GCSM}^{i\times{j}}$ is defined as
\begin{align}\label{eq:gcsm}
\begin{split}
{k_\text{GCSM}^{i\times{j}}(\tau)}
=&\mathcal{F}_{s\rightarrow \tau}^{-1}\left[\frac{1}{2}\left({\freq{k}_{{\gcsm}}^{\itimej}({\vs})} + {\freq{k}_{{\gcsm}}^{\itimej}({-\vs})}\right)\right](\tau)\\
=&{w_{ij}a_{ij}}\exp\left(-\frac{\pi^{2}}{2}{(2\tau-{\vtime}_{ij})\tra{{\Var}_{ij}}(2\tau-{\vtime}_{ij})}\right)\\
&\times\cos\left({\pi}\left( {(2\tau-{\vtime}_{ij})\tra{\vmu}_{ij}}-{{\vphase}_{ij}}\right)\right)
\end{split}
\end{align}
Here
\begin{itemize}
	\item  {$w_{ij}=\sqrt{w_{i}w_{j}}$} is the cross weight;	
	\item 
	{${a}_{ij}={\left|\frac{\sqrt{4{\Var}_{i}{\Var}_{j}}}{{\Var}_{i}+{\Var}_{j}}\right|}^{\frac{1}{2}}\\
		\phantom{===}\times\exp\left(-\frac{1}{4}{({\vmu}_{i}-{\vmu}_{j})\tra}({{\Var}_{i}+{\Var}_{j}})^{-1}({\vmu}_{i}-{\vmu}_{j})\right)$} is the cross amplitude;
	\item  {$\vmu_{ij}=\frac {{\Var}_{i}{\vmu}_{j}+{\Var}_{j}{\vmu}_{i}}{{{\Var}_{i}+{\Var}_{j}}}$} is the cross mean;
	\item  {${{\Var}}_{ij}=\frac{{2{{\Var}_{i}{\Var}_{j}}}}{{{\Var}_{i}+{\Var}_{j}}}$} is the cross covariance;
	\item {${\vtime}_{ij}={\vtime}_{i}-{\vtime}_{j}$} is the cross time delay; and 
	\item  {${\vphase}_{ij}={\vphase}_{i}-{\vphase}_{j}$} is the cross phase delay.
\end{itemize}

\begin{figure*}[h!]
\centering
\renewcommand{\tabcolsep}{0.5mm}
\begin{tabular}{p{0.5mm}*{4}{c}}
  & \includegraphics[width=0.5\columnwidth]{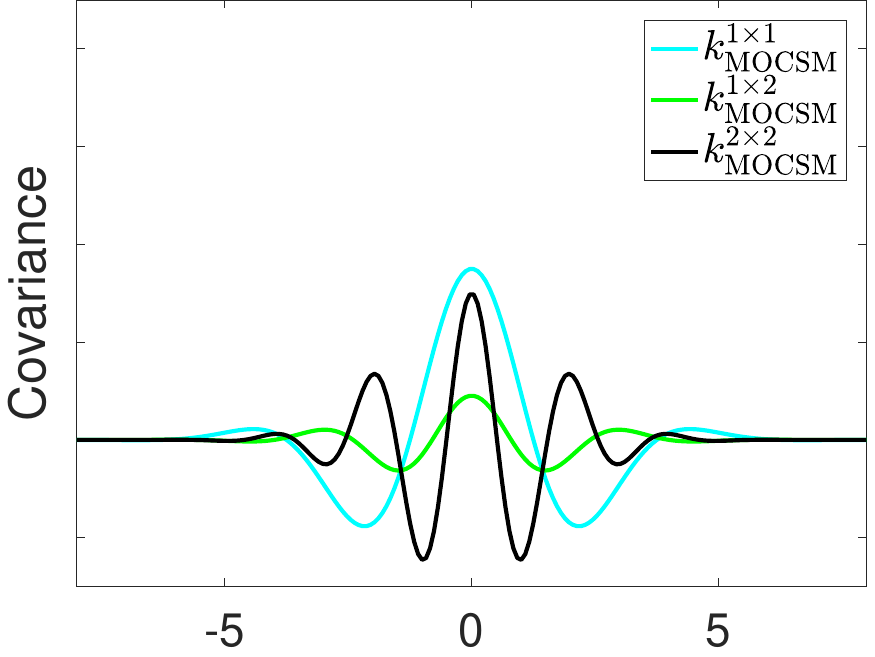} 
 & \includegraphics[width=0.5\columnwidth]{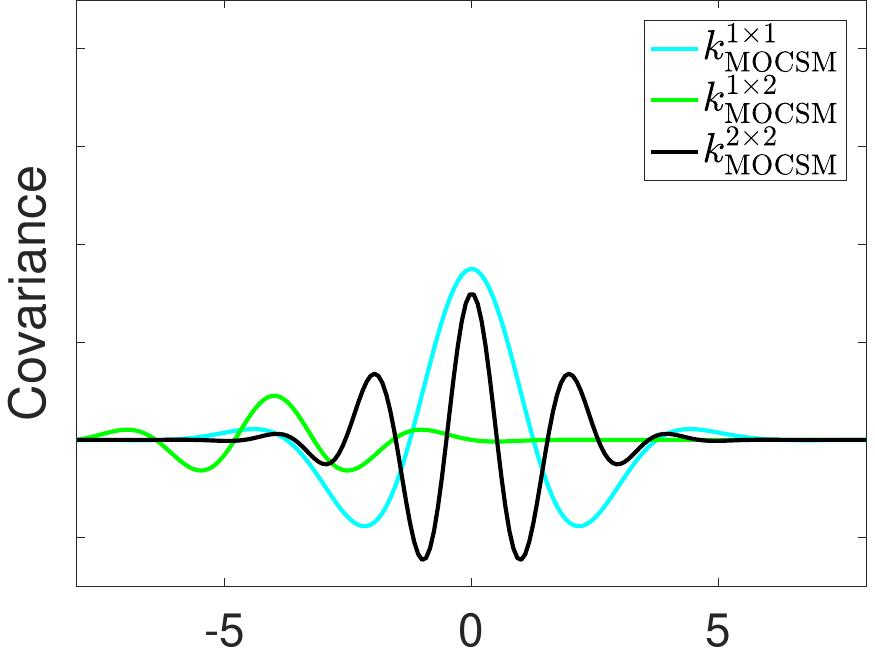}
  & \includegraphics[width=0.5\columnwidth]{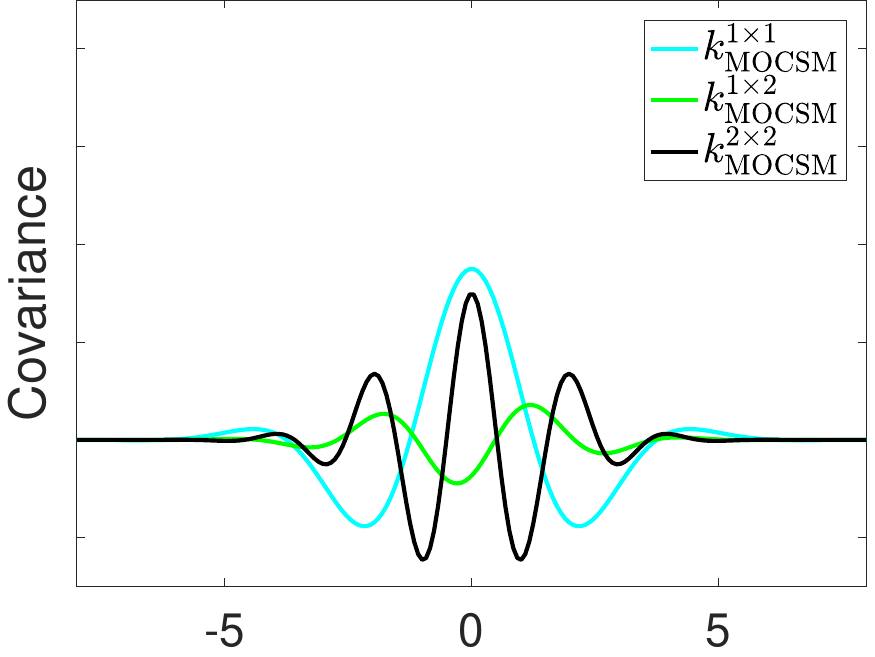}
 & \includegraphics[width=0.5\columnwidth]{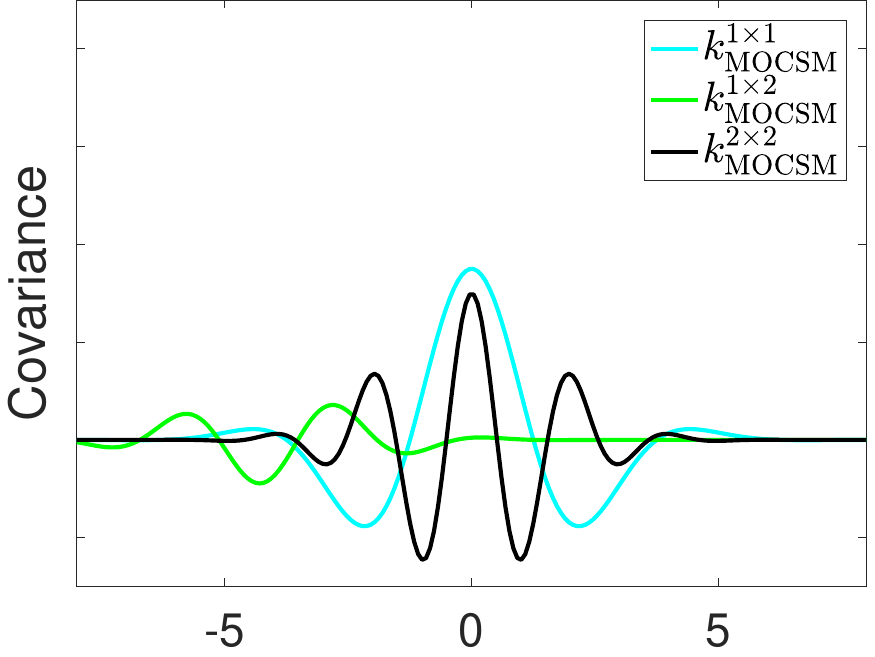}\\
  & \includegraphics[width=0.5\columnwidth]{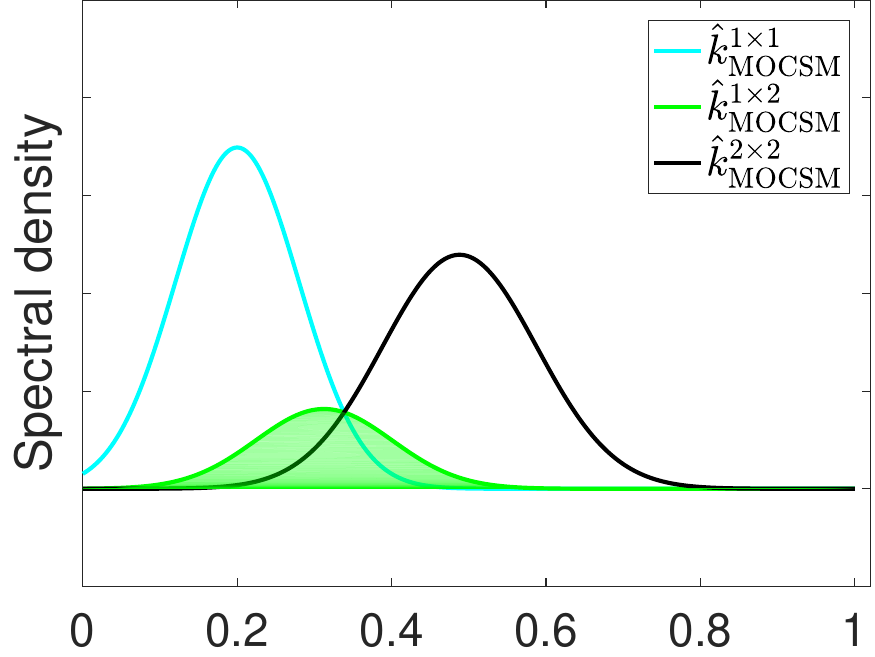} 
 & \includegraphics[width=0.5\columnwidth]{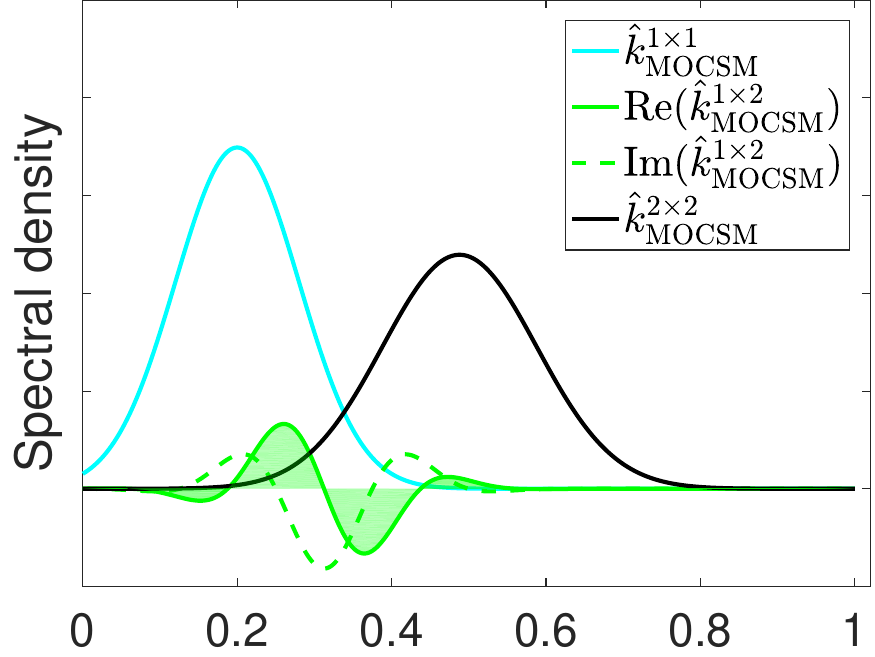}
  & \includegraphics[width=0.5\columnwidth]{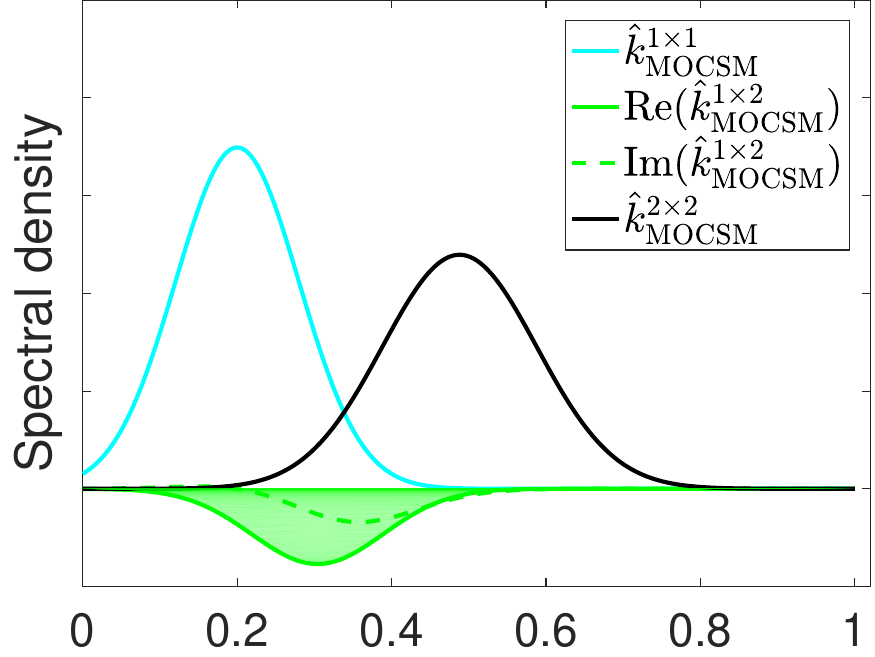} 
 & \includegraphics[width=0.5\columnwidth]{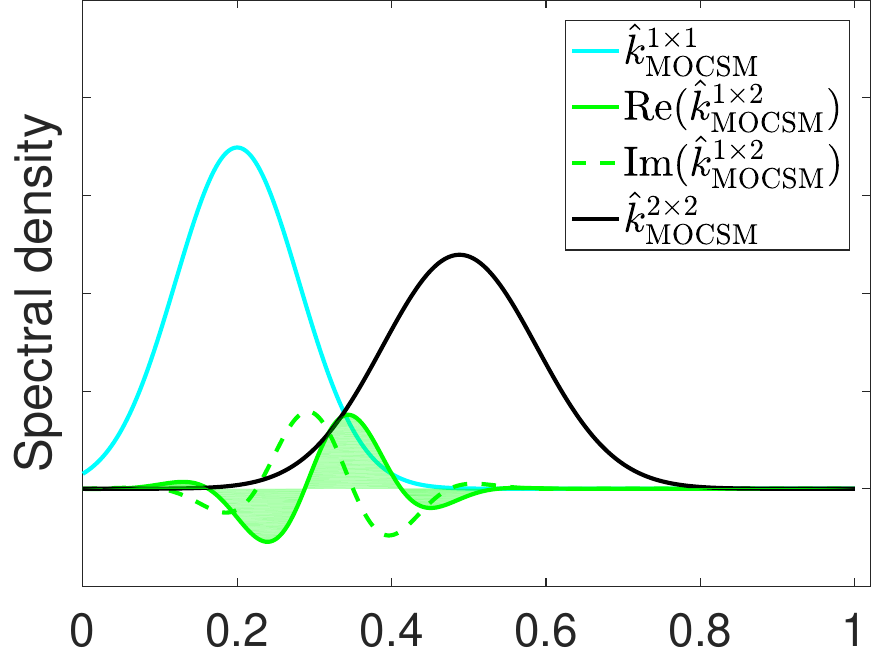}\\
  & {(a) $\vtime_{ij}=0, \vphase_{ij}=0$} & {(b) $\vtime_{ij}\neq0, \vphase_{ij}=0$} & {(c) $\vtime_{ij}=0, \vphase_{ij}\neq0$} & {(d) $\vtime_{ij}\neq0, \vphase_{ij}\neq0$}
 \end{tabular}
	\caption{Illustration and interpretation of cross convolution between channels in MOCSM. The green shadow region describes intersection of two channels modeled by Gaussians (respectively in black and cyan), which represents cross correlation (the shared frequency range) between channels. Here we model the shared frequency range as a Gaussian (in green and green shadow). First row: cross convolution of MOCSM with zero time and phase delay, non-zero time delay and zero phase delay, zero time delay and non-zero phase delay, non-zero time and phase delay. Second row: corresponding spectral density of the first row. Operators $\mathrm{Re}$ and $\mathrm{Im}$ respectively return real and imaginary parts. }\label{fig:gcsm-illustration}
\end{figure*}

In the next section we further extend the spectral mixture kernel by modeling dependencies between tasks in a Multi-Output Gaussian Process (MOGP) scenario. 

\subsection{Multi-output convolution spectral mixture kernel}
 The above described approach based on convolution decomposition of a generalized (complex valued) spectral density function, can be directly used in a multi-output setting, 
by considering  $\tau=\vx^i - \vx^j$ in the formula for ${k_\text{GCSM}^{i\times{j}}(\tau)}$, where $x^i$ and $x^j$ are inputs from the $i$-th and $j$-th channel, respectively.   We call the resulting kernel Multi-Output Convolution Spectral Mixture (MOCSM) kernel.
 
In a multi-output setting, the formulation for the cross spectral density  ${\freq{k}_\text{MOCSM}^{i\times{j}}(\vs)}$ is as that  for  ${\freq{k}_\text{GCSM}^{i\times{j}}(\vs)}$ by taking  $i,j\in\{1,...,M\}$ to be channels (instead of components). From Equations (\ref{eq:gcsm_spec}), (\ref{eq:gcsm}), and Figure \ref{fig:gcsm-illustration}, we can interpretate the correlation between channels as: there are intersection between channels described by Gaussians, their shared frequency range encodes their correlation in the frequency domain, which can be modeled as a Gaussian by cross convolution.  In order for ${{k}_\text{MOCSM}^{i\times{j}}({\tau})}$ to be a kernel,  it should be positive semi-definite, which is true if and only if its spectral density ${\freq{k}_\text{MOCSM}^{i\times{j}}(\vs)}$ is positive semi-definite \cite{Bochner2016,Stein}.  
Therefore we demonstrate the latter condition. Given any finite set of non-zero vectors $[{\vz}_{1}, ..., {\vz}_{M}]\tra\in\mathbb{C}^{M\times{P}}$ with complex values entries for $M$ outputs, ${\vs}\in{\bbR}^{{P}}$, we have
\begin{align}\label{eq:psd-half}
\begin{split}
\sum_{i,j=1}^{M}\left({\vz}_{i}{\freq{k}_\text{MOCSM}^{i\times{j}}(\vs)}{{\vz}_{j}^{\dagger}}\right)
&=\sum_{i,j=1}^{M}\left({\vz}_{i}\left({\hat{g}_{\text{GCSM}i}({\vs})}\cdot\overline{\hat{g}_{\text{GCSM}j}}({\vs})\right){{\vz}_{j}^{\dagger}}\right)\\
&=\sum_{i,j=1}^{M}{\,}\left({\vz}_{i}{\hat{g}_{\text{GCSM}i}({\vs})}\right)\cdot\left(\overline{{\vz}_{j}\hat{g}_{\text{GCSM}j}}({\vs})\right)\\
&=\Big{|}\sum_{i=1}^{M}{{\vz}_{i}\freq{g}_{{\gcsm}i}({\vs})}\Big{|}^{2}\geq{0}
\end{split}
\end{align}
where ${\vz}_{i}^{\dagger}$ denotes the conjugate transpose of ${\vz}_{i}$.  So ${\freq{k}_\text{MOCSM}^{i\times{j}}(\vs)}$ satisfies the positive definite condition.

From this result it follows that also a $Q$-components, multi-dimensional input MOCSM kernel is positive definite. The resulting MOCSM kernel has the following form:
\begin{align}\label{eq:MOCSM}
\begin{split}
&k^{i \times j} _\text{MOCSM}(\tau)\\
&=\sum_{q=1}^{Q}{c_{ij}^{(q)}}\exp\left(-\frac{\pi^{2}}{2}{\left(2\tau-{\vtime}_{ij}^{(q)}\right)\tra{{\Var}_{ij}^{(q)}}\left(2\tau-{\vtime}_{ij}^{(q)}\right)}\right)\\
&\phantom{==}\times\cos\left({\pi}\left( {(2\tau-{\vtime}_{ij}^{(q)})\tra{\vmu}_{ij}^{(q)}}-{{\vphase}_{ij}^{(q)}}\right)\right),
\end{split}
\end{align}
where the $q$-th cross contribution $c_{ij}^{(q)}={w_{ij}^{(q)}}{a_{ij}^{(q)}}$ incorporates the $q$-th cross weight and the $q$-th cross amplitude between $i$-th and $j$-th channels.
Here 
\begin{itemize}	
	\item  {$w_{ij}^{(q)}=\sqrt{{w_{i}^{(q)}}{w_{j}^{(q)}}}$} is $q$-th cross weight;
	\item 
	{${a}_{ij}^{(q)}={\left|\frac{\sqrt{4{\Var}_{i}^{(q)}{\Var}_{j}^{(q)}}}{{\Var}_{i}^{(q)}+{\Var}_{j}^{(q)}}\right|}^{\frac{1}{2}}\\
	\phantom{===.}\times\exp\Big(-\frac{1}{4}{\left({\vmu}_{i}^{(q)}-{\vmu}_{j}^{(q)}\right)\tra}({{\Var}_{i}^{(q)}+{\Var}_{j}^{(q)}})^{-1}\\
	\phantom{========}\times\left({\vmu}_{i}^{(q)}-{\vmu}_{j}^{(q)}\right)\Big)$} is the $q$-th cross amplitude;
	\item  {$\vmu_{ij}^{(q)}=\frac {{\Var}_{i}^{(q)}{\vmu}_{j}^{(q)}+{\Var}_{j}^{(q)}{\vmu}_{i}^{(q)}}{{{\Var}_{i}^{(q)}+{\Var}_{j}^{(q)}}}$} is the $q$-th cross mean;
	\item  {${{\Var}}_{ij}^{(q)}=\frac{{2{{\Var}_{i}^{(q)}{\Var}_{j}^{(q)}}}}{{{\Var}_{i}^{(q)}+{\Var}_{j}^{(q)}}}$} is the $q$-th cross covariance;
	\item  {${\vtime}_{ij}^{(q)}={\vtime}_{i}^{(q)}-{\vtime}_{j}^{(q)}$} is the $q$-th cross time delay; and 
	\item  {${\vphase}_{ij}^{(q)}={\vphase}_{i}^{(q)}-{\vphase}_{j}^{(q)}$} is the  $q$-th cross phase delay.
\end{itemize}

In the above Equations, $\Sigma_{i}$, $\boldsymbol{\mu}_{i}$, $\boldsymbol{\theta}_{i}$, $\boldsymbol{\phi}_{i}$ have the dimension $P$ as in Equation (\ref{eq:smp}). 
In our MOCSM kernel, one can also employ angular frequencies instead of ordinary frequencies in the convolution and the inverse Fourier transform. 
Next, we analyze and compare the MOCSM kernel with the MOSM and the SM kernel.

\subsection{Comparison between MOCSM and MOSM}\label{sec:diff}
Here for simplicity of exposition we compare MOCSM and MOSM employing angular frequencies with only one component. We consider the parameters described in the last section and use superscript to respectively denote the different terms in MOSM and MOCSM. 
Recall that MOSM and MOCSM are obtained using different approaches: MOSM employs a result on complex-valued matrix factorization to construct a (complex valued) squared exponential function as a cross spectral density, while MOCSM uses the convolution theorem.

\begin{table*}[!ht]
	\caption{Summary of differences between MOCSM and MOSM in terms of  methods, approach,  cross weight, cross amplitude, cross phase, and  single output case ($M=1$).} \label{tab:diff}
	\begin{center}
	\newcolumntype{C}{ >{\centering\arraybackslash}m{4cm}}
	\newcolumntype{D}{ >{\centering\arraybackslash}m{4.0cm}}
	\newcolumntype{E}{ >{\centering\arraybackslash}m{3.2cm}}
	\newcolumntype{F}{ >{\centering\arraybackslash}m{5cm}}	
	\renewcommand*{\arraystretch}{1.2}
	\begin{tabular}{CFDE}
		\toprule
		{Difference} & {MOCSM} & {MOSM} & {SM} \\
		\midrule
		{Methods} & {Convolution Theorem} & {Complex-valued matrix decomposition} & Bochner's Theorem \\		
		
		{Approach} & {Modeling the intersection of channels as a Gaussian spectral density by using convolution} & {Modeling cross spectral density as a complex-valued positive-definite matrix} & {Modeling spectral density as a mixture of Gaussians} \\		
        
		cross weight of covariance \newline and spectral density & ${\sqrt{w_{i}w_{j}}}$ &  $w_{i}w_{j}$ & not available \\
        
        cross amplitude of covariance & ${\left|\frac{\sqrt{4{\Var}_{i}{\Var}_{j}}}{{\Var}_{i}+{\Var}_{j}}\right|}^{\frac{1}{2}}$ &  $(2\pi)^{\frac{n}{2}}{|\Sigma_{ij}|^{1/2}}$ & not available \\
        
		cross amplitude of spectral density & $\frac{1}{\sqrt{(2\pi)^{P}{|\Sigma_{ij}|}}}{\left|\frac{\sqrt{4{\Var}_{i}{\Var}_{j}}}{{\Var}_{i}+{\Var}_{j}}\right|}^{\frac{1}{2}}$ &  1 & not available \\
        
		cross phase &  $\vphase_{ij}=\vphase_{i}-\vphase_{j}$, vector $\vphase_{i}\in\bbR^{P}$  &  $\phase_{ij}=\phase_{i}-\phase_{j}$, scalar $\phase_{i}\in\bbR$  & not available  \\
        
		single channel weight of covariance and spectral density &  $w_{i}$  &  $w_{i}^{2}$  & $w_{i}$  \\
		
		single channel amplitude of covariance  &  1 &  $(2\pi)^{\frac{n}{2}}{|\Sigma_{i}|^{1/2}}$  & 1  \\
		
		single channel amplitude of spectral density &  $\frac{1}{\sqrt {(2\pi )^{P}|{\Sigma_{i}}|}}$  &  $1$  & $\frac{1}{\sqrt {(2\pi )^{P}|{\Sigma_{i}}|}}$  \\		
        
		single output case (with $M=1$) & ${w_i}k_{\sm{i}}$ & ${w_{i}^{2}(2\pi)^{\frac{n}{2}}{|\Sigma_{i}|^{1/2}}}k_{\sm{i}}$ & ${w_i}k_{\sm{i}}$ \\
		\bottomrule
	\end{tabular}
\end{center}
\end{table*}

Specifically, for MOSM, in the spectral domain we have:
    \begin{align}\label{eq:mosm_spec}
    \begin{split}
    \freq{k}&^{\mosm}_{ij}(\vs)\\
    &=w_{ij}^{\mosm}{\exp\left(-\frac{1}{4}\left(\vmu_{i}-\vmu_{j})^{\top}(\Sigma_{i}+\Sigma_{j})^{-1}(\vmu_{i}-\vmu_{j}\right)\right)}\\
    &\phantom{=}\times\exp\left(-\frac{1}{2}\left(\vs-\vmu_{ij})^{\top}\Sigma_{ij}^{-1}(\vs-\vmu_{ij}\right)\right)\\
    &\phantom{=}\times\exp\left(\imagi\left(\vtime_{ij}^{\top}\vs+\phi_{ij}^{\mosm}\right)\right)
    \end{split}
    \end{align}
    
    which yields the  kernel
    
    \begin{align}\label{eq:mosm_one}
    \begin{split}
    {k}^{\mosm}_{ij}(\tau)=&w_{ij}^{\mosm}(2\pi)^{\frac{n}{2}}{|\Sigma_{ij}|^{1/2}}\\&\times\exp\left(-\frac{1}{4}\left(\vmu_{i}-\vmu_{j})^{\top}(\Sigma_{i}+\Sigma_{j})^{-1}(\vmu_{i}-\vmu_{j}\right)\right)\\
    &\times\exp\left(-\frac{1}{2}\left(\tau+\vtime_{ij}\right)^{\top}\Sigma_{ij}\left(\tau+\vtime_{ij}\right)\right)\\&\times\cos\left(\left(\tau+\vtime_{ij}\right)^{\top}\vmu_{ij}+\phi_{ij}^{\mosm}\right)
    \end{split}%\\
\end{align}
 Here
    \begin{itemize}
	\item  	$w_{ij}^{\mosm}=w_{i}w_{j}$ is cross weight in MOSM;
	\item {$a_{ij}^{\mosm}=(2\pi)^{\frac{n}{2}}{|\Sigma_{ij}|^{1/2}}\\
	\phantom{=====}
	\times\exp\left(-\frac{1}{4}\left(\vmu_{i}-\vmu_{j})^{\top}(\Sigma_{i}+\Sigma_{j})^{-1}(\vmu_{i}-\vmu_{j}\right)\right)$ represents the cross amplitude in MOSM;}
	\item  {${\phi}_{ij}^{\mosm}={\phi}_{i}-{\phi}_{j}$} is the cross phase delay (a scalar).%; and
\end{itemize}
On the other hand, for MOCSM, in the spectral domain we have:
\begin{align}
    \begin{split}
    \freq{k}&^{\mocsm}_{i}(\vs)\\
    &=w_{i}\uptext{\frac{1}{\sqrt{(2\pi)^{P}|\Sigma_{i}|}}\exp\left(-\frac{1}{2}\left(\vs-\vmu_{i})^{\top}\Sigma_{i}^{-1}(\vs-\vmu_{i}\right)\right)}^{\text{A Gaussian distribution with integral equal to 1}}\\
    &\times\exp\left(-2\pi\imagi\left(\vtime_{i}^{\top}\vs+\vphase_{i}\right)\right)
    \end{split}
    \end{align}
From Equation (\ref{eq:gcsm_spec}), the cross spectral density of MOCSM is 
    \begin{align}\label{eq:mocsm_one}
    \begin{split}
    \freq{k}^{\mocsm}_{ij}(\vs)=
    &w_{ij}^{\mocsm}\frac{1}{\sqrt{(2\pi)^{P}{|\Sigma_{ij}|}}}{\left|\frac{\sqrt{4{\Var}_{i}{\Var}_{j}}}{{\Var}_{i}+{\Var}_{j}}\right|}^{\frac{1}{2}}\\&\times\exp\left(-\frac{1}{4}\left(\vmu_{i}-\vmu_{j})^{\top}(\Sigma_{i}+\Sigma_{j})^{-1}(\vmu_{i}-\vmu_{j}\right)\right)\\&\times\uptext{\exp\left(-\frac{1}{2}\left(\vs-\vmu_{ij})^{\top}\Sigma_{ij}^{-1}(\vs-\vmu_{ij}\right)\right)}^\text{exponential part of the Gaussian cross spectral density}\\&\times\exp\left(-\pi\imagi\left(\vtime_{ij}^{\top}\vs+\vphase_{ij}^{\mocsm}\right)\right).
    \end{split}
\end{align}
The cross component of MOCSM is given in Equation (\ref{eq:gcsm}). Here  $a_{ij}^{\mocsm}=a_{ij}$ 
is the amplitude of a Gaussian cross spectral density from a product of two Gaussian spectral densities. For MOCSM the resulting kernel (in the considered single-component setting) is Equation (\ref{eq:MOCSM}) with $Q=1$.
    For MOCSM we have
    \begin{itemize}
	\item  	$w_{ij}^{\mocsm}=\sqrt{w_{i}w_{j}}$ is cross weight in MOCSM;
 	\item  {${\vphase}_{ij}^{\mocsm}={\vphase}_{i}-{\vphase}_{j}$} is the cross phase delay (a vector).	
\end{itemize}

\begin{figure*}[!ht]
	\centering 	
\begin{subfigure}[t]{0.245\linewidth}
	\centering
	\includegraphics[width=1.1\linewidth]{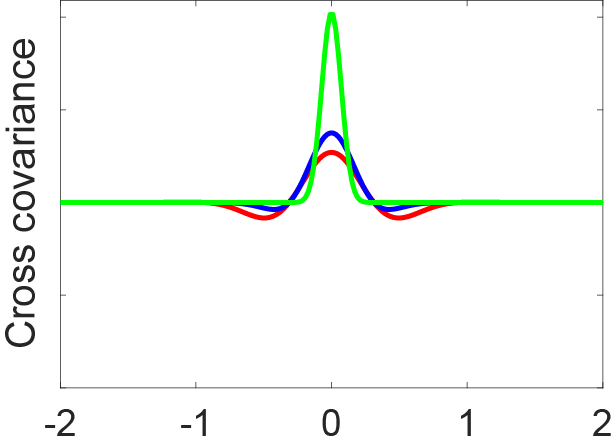}
	\caption{}      
\end{subfigure} 
\begin{subfigure}[t]{0.245\linewidth}
	\centering
	\includegraphics[width=1.1\linewidth]{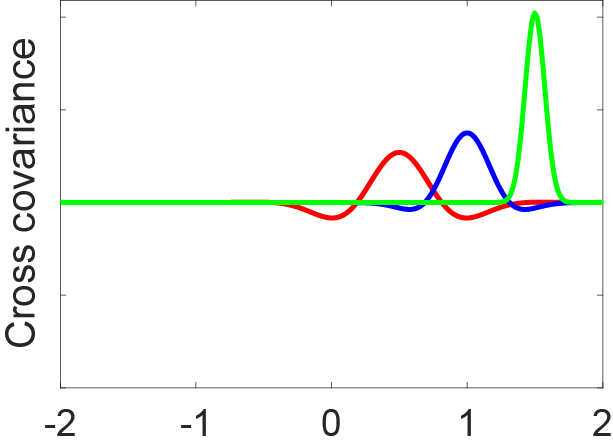}
	\caption{}     
\end{subfigure}
\begin{subfigure}[t]{0.245\linewidth}
	\centering
	\includegraphics[width=1.1\linewidth]{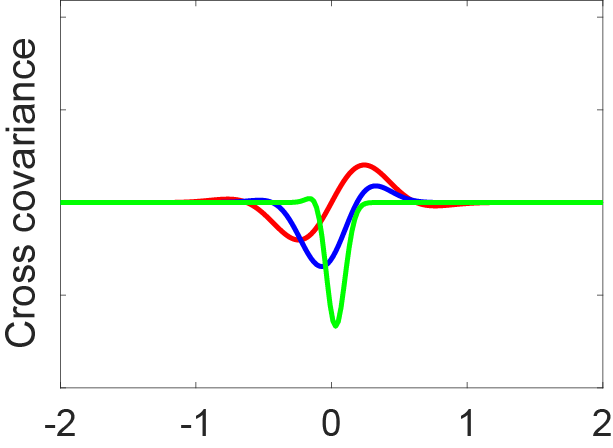}
	\caption{}    
\end{subfigure}
\begin{subfigure}[t]{0.245\linewidth}
	\centering
	\includegraphics[width=1.1\linewidth]{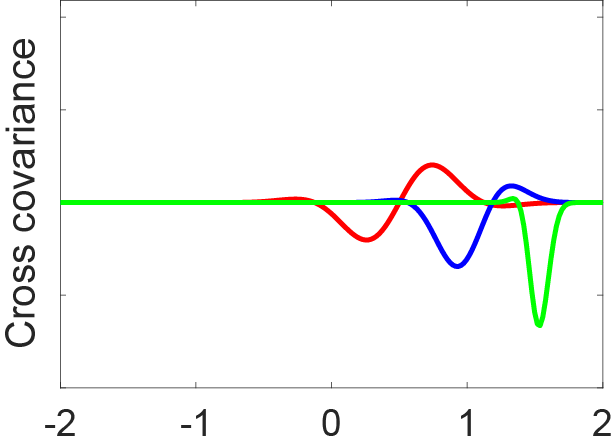}
	\caption{} 
\end{subfigure}

\begin{subfigure}[t]{0.245\linewidth}
	\centering
	\includegraphics[width=1.1\linewidth]{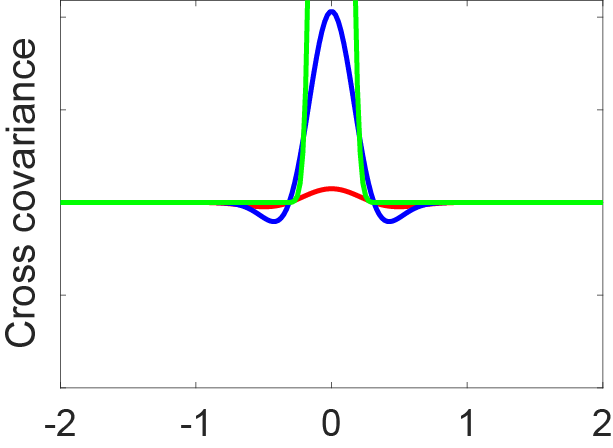}
	\caption{}    
\end{subfigure} 
\begin{subfigure}[t]{0.245\linewidth}
	\centering
	\includegraphics[width=1.1\linewidth]{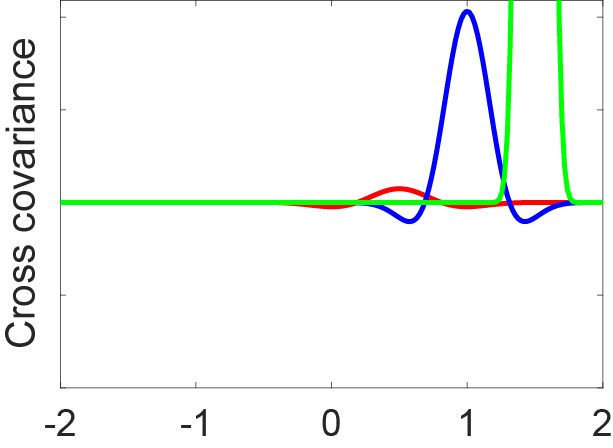}
	\caption{} 
\end{subfigure}
\begin{subfigure}[t]{0.245\linewidth}
	\centering
	\includegraphics[width=1.1\linewidth]{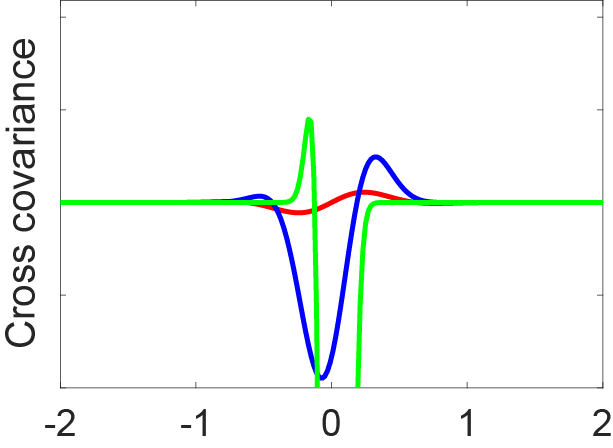}
	\caption{}
\end{subfigure}
\begin{subfigure}[t]{0.245\linewidth}
	\centering
	\includegraphics[width=1.1\linewidth]{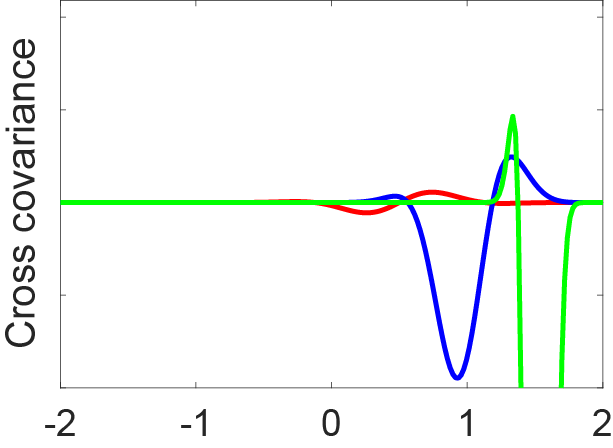}
	\caption{}
\end{subfigure}

\caption{Cross covariances of $k_\text{MOCSM}(\tau)$ (first row) and $k_\text{MOSM}(\tau)$ (second row) with four channels. The columns show the channel dependencies with respect to zero time and phase delay, non-zero time delay and zero phase delay, zero time delay and non-zero phase delay, non-zero time and phase delay cross convolution. We only show three cross covariances: cov(channel$^{(1)}$, channel$^{(2)}$) with red solid line, cov(channel$^{(2)}$, channel$^{(3)}$) with blue solid line, cov(channel$^{(3)}$, channel$^{(4)}$) with green solid line.}
\label{fig:comp}
\end{figure*}

MOCSM, MOSM, and other SM based multi-output kernels have similar exponential and similar cosine terms, but really different spectral density and kernel forms: cross weight  ($w_{ij}^{\mosm}\neq{w_{ij}^{\mocsm}}$), cross amplitude ($a_{ij}^{\mosm}\neq{a_{ij}^{\mocsm}}$), and cross phase delay ($\phi_{ij}^{\mosm}\neq{\vphase_{ij}^{\mocsm}}$) terms (see also Table \ref{tab:diff}, Figure \ref{fig:comp}, and Figure  \ref{fig:comp}).

\begin{figure*}[!ht]
	\centering 	
\begin{subfigure}[t]{0.245\linewidth}
	\centering
	\includegraphics[width=1.05\linewidth]{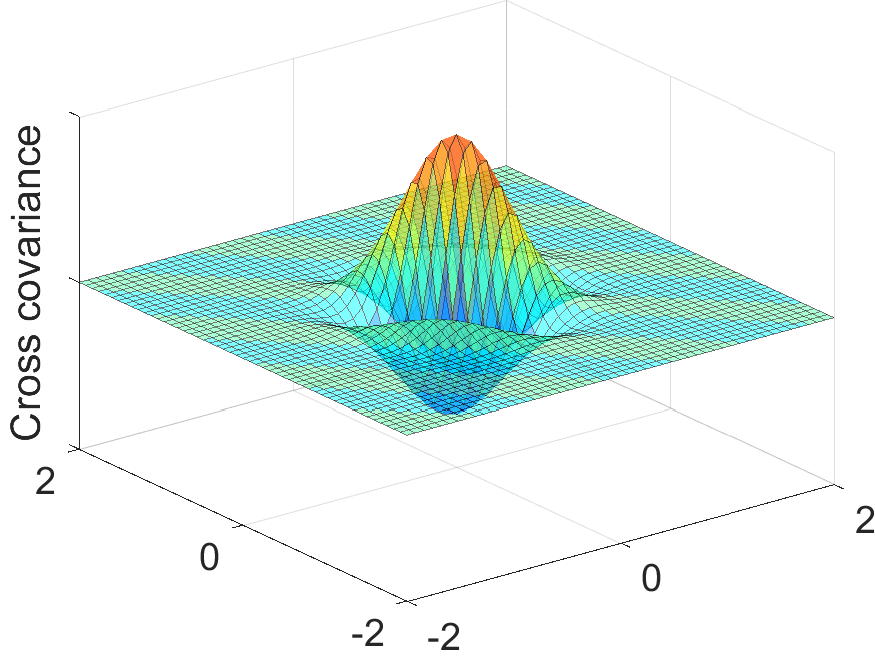}
	\caption{MOCSM ($\vtime_{ij}=\mathbf{0}, \vphase_{ij}=\mathbf{0}$)\\  and MOSM ($\vtime_{ij}=\mathbf{0}, \phi_{ij}=0$).}
\end{subfigure}
\begin{subfigure}[t]{0.245\linewidth}
	\centering
	\includegraphics[width=1.05\linewidth]{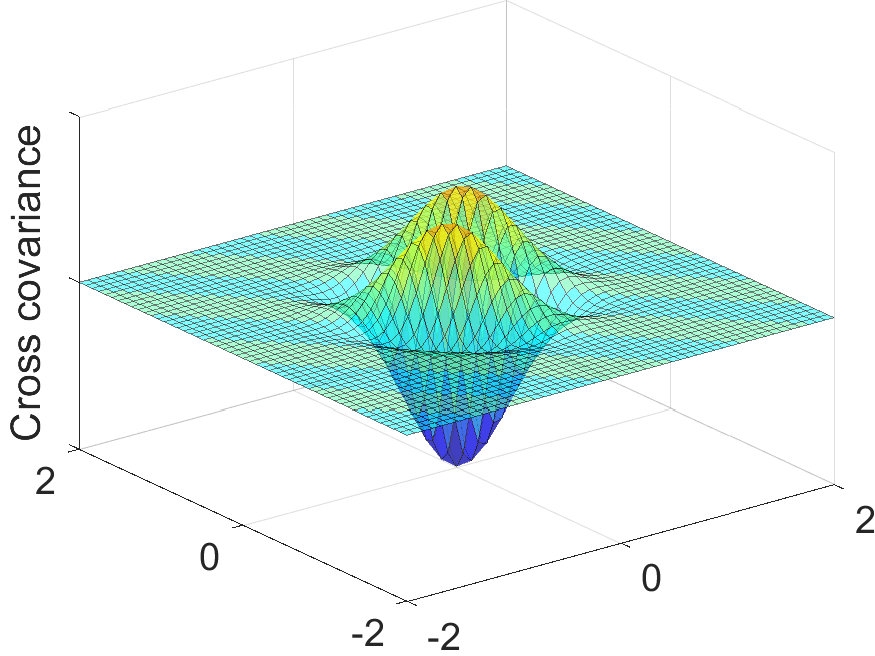}
	\caption{MOCSM ($\vtime_{ij}=\mathbf{0}, \vphase_{ij}\neq\mathbf{0}$\\ with $\phi_{ij, 1}=\phi_{ij, 2}$) and MOSM\\ ($\vtime_{ij}=\mathbf{0}, \phi_{ij}\neq0$).}
\end{subfigure}
\begin{subfigure}[t]{0.245\linewidth}
	\centering
	\includegraphics[width=1.05\linewidth]{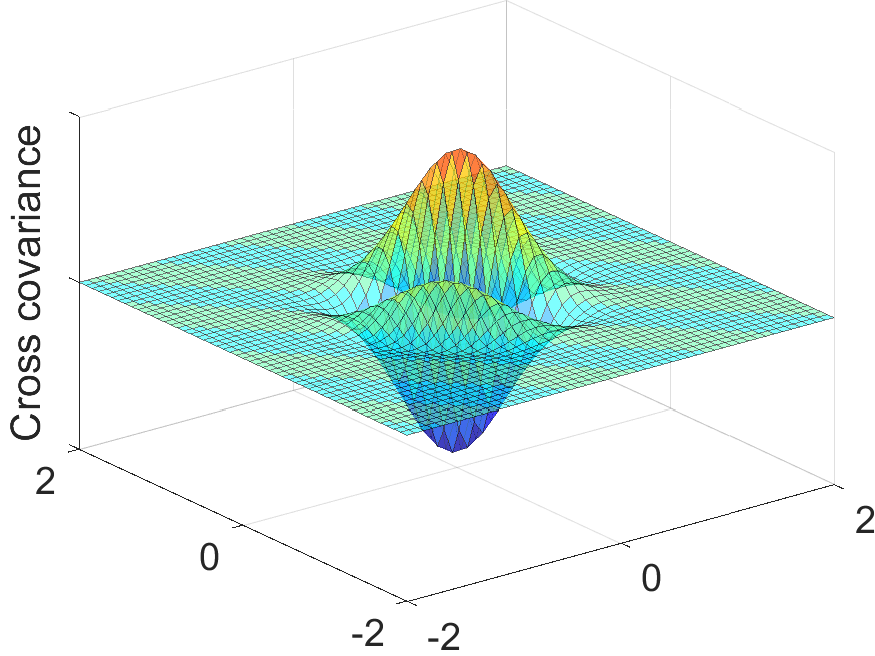}
	\caption{MOCSM ($\vtime_{ij}=\mathbf{0}, \vphase_{ij}\neq\mathbf{0}$ \\ with $\phi_{ij, 1}\neq\phi_{ij, 2}$).}
\end{subfigure}
\begin{subfigure}[t]{0.245\linewidth}
	\centering
	\includegraphics[width=1.05\linewidth]{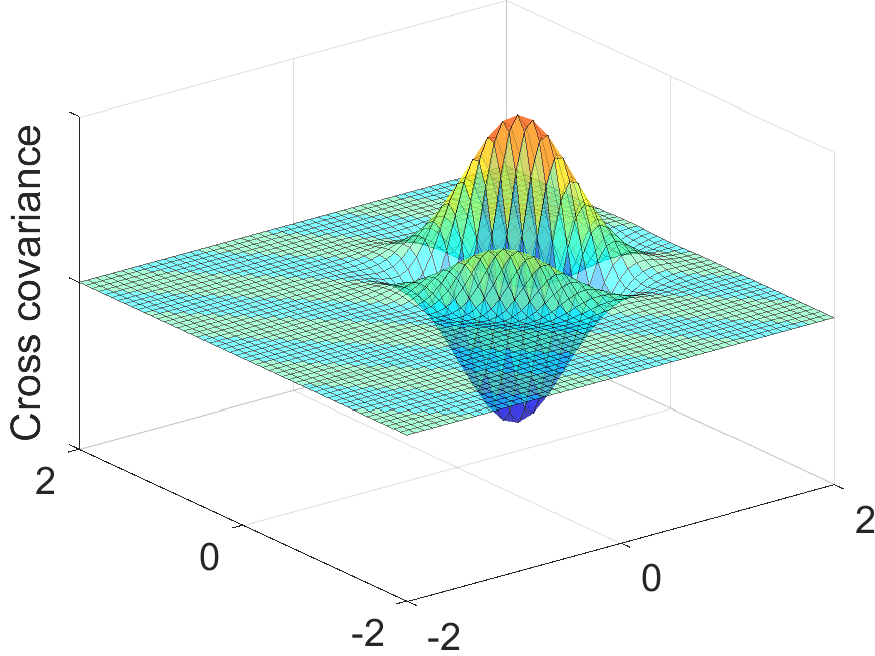}
	\caption{MOCSM ($\vtime_{ij}\neq\mathbf{0}, \vphase_{ij}\neq\mathbf{0}$ \\
	with $\phi_{ij, 1}\neq\phi_{ij, 2}$).}
\end{subfigure}
\caption{Cross covariances of two dimensional inputs for $k_\text{MOCSM}(\tau)$ and $k_\text{MOSM}(\tau)$ with two channels. The first two subplots (a) and (b) show channel dependencies with respect to zero time and phase delay, zero time delay and non-zero phase delay for both MOSM and  MOCSM. 
The last two subplots (c) and (d) show zero time delay and non-zero phase delay, non-zero time and non-zero phase delay for MOCSM (each input dimension has a different phase delay $\phi_{ij, 1}\neq\phi_{ij, 2}$).}
\label{fig:comp2}
\end{figure*}

To illustrate the consequences of such discrepancy, consider a setting with two channels $i$ and $j$. Suppose the correct weights $(w_i, w_j)$ and variances $(\sigma_i^{2}, \sigma_j^{2})$ of channels $i$ and $j$ are close to each other. 
In MOSM the cross weight term will amplify the cross correlation between the two channels when $|w_i|>1$, $|w_j|>1$, $|\sigma_i|>1$, and $|\sigma_j|>1$, since MOSM contains squared weights (see also Table \ref{tab:diff}). 
In Figure \ref{fig:comp2} we illustrate this phenomenon for 4 channels with weights  $\boldsymbol{w}=(0.5, 0.6, 2.0, 2.1)$ and variances   $\boldsymbol{\sigma}^{2}=(0.4, 0.5, 2.0, 2.1)$, respectively. The plots show that the cross weights in MOSM may be amplified or may shrunk because of the square used in the cross channel weights. In particular, in MOSM (second row), cov(channel$^{(1)}$, channel$^{(2)}$) (in red solid line) is smaller than that of MOCSM (in first row), while cov(channel$^{(2)}$, channel$^{(3)}$) (in blue solid line) is larger than that of MOCSM. Notably cov(channel$^{(3)}$, channel$^{(4)}$) (in green solid line) is much larger in MOSM than in MOCSM. 

We now illustrate the differences between MOSM and MOCSM in a setting with two channels and a two-dimensional input (see Figure \ref{fig:comp}). Unlike in MOSM, where all input dimensions are enforced to have equal phase determined by $\phi_{ij}$, MOCSM allows to model phase delay on each input dimension, that is $\phi_{ij,1}$, $\phi_{ij,2}$.
 
Finally, we compare the  MOSM and MOCSM kernels with the SM one. Recall that SM kernel is

\begin{align}\label{eq:sm}
    \begin{split}
    {k}^{\sm}_{i}(\tau)=&\uptext{w_{i}}^{\text{signal variance}}\exp\bigg(-\frac{1}{2}\tau^{\top}\uptext{\Sigma_{i}}^{\text{length-scale}}\tau\bigg)\cos\left(\tau^{\top}\vmu_{i}\right)
    \end{split}      
\end{align}

For a single channel the MOCSM kernel becomes
\begin{align}
    \begin{split}
    {k}^{\mocsm}_{i}(\tau)=&w_{i}\exp\left(-\frac{1}{2}\tau^{\top}\Sigma_{i}\tau\right)\cos\left(\tau^{\top}\vmu_{i}\right),
    \end{split}
\end{align}
which is equal to the SM kernel.
However, for a single channel, the MOSM kernel becomes

\begin{align}
 \begin{split}\label{eq:reduce_mosm}
    {k}^{\mosm}_{i}(\tau)=&\uptext{w_{i}^{2}(2\pi)^{\frac{n}{2}}{|\Sigma_{i}|^{1/2}}}^{\text{squared signal variance mixed with length-scale}}\\&\phantom{===}\times\exp\left(-\frac{1}{2}\tau^{\top}\Sigma_{i}\tau\right)\cos\left(\tau^{\top}\vmu_{i}\right).
    \end{split}  
\end{align}

Thus MOSM cannot capture single channel patterns as SM does because the variance (a length-scale parameter) contributes also to the the signal variance. The proposed MOCSM also has a better generalization ability than MOSM. Overall, MOCSM are different from MOSM in terms of motivation, interpretation, designing method, and mathematical description. 

\section{Related work}\label{sec:rw}
There is an abundant literature on GPs related to MOGPs \cite{Parra2017,bonilla2008multi, Melkumyan2011,Alvarez2009,Wilson2011, Ulrich2015a,Duvenaud2013}. Here we mainly focus on MOGPs methods based on spectral mixture kernels, because of their expressiveness and recent use in MOGPs. Since the introduction of SM kernels \cite{Wilson2014, Wilson2013}, various MOGP methods have been introduced \cite{Wilson2014, Duvenaud2013, Flaxman2015, Oliva2016, Jang2017}. The first such MOGP kernel, based on the LMC framework introduced in \cite{Wilson2011} was defined as
$${K}_\text{SM-LMC}=\sum_{i=1}^{Q}{B_{i}}\otimes{K_{\text{SM}{i}}}$$ 
Here ${B_{i}}$  encodes cross weights to represent channel correlations
and involves a linear combination of components. 
The CSM kernel \cite{Ulrich2015a} improved the expressiveness of SM-LMC by introducing a cross phase spectrum and was also defined within the LMC framework as 
$${K}_\text{CSM}=\sum_{i=1}^{Q}{B_{i}}{k_{\text{SG}i}}(\tau;{\Theta}_{i})$$
where ${k_{\text{SG}i}}(\tau;{\Theta}_{i})$ is phasor notation of the spectral Gaussian kernel. However the kernels ${k_{\text{SG}i}}(\tau;{\Theta}_{i})$ used in the CSM are only phase dependent,  but not time dependent. The more recent MOSM kernel \cite{Parra2017} provided a principled framework to construct multivariate covariance functions with a better interpretation of cross correlations between channels. SM-LMC and CSM are instances of MOSM. However, MOSM cannot reduce to the SM kernel when using only one channel and in certain situations produces inaccurate channel correlations, for example, when patterns between channels are close to each other (see Table \ref{tab:diff}). Note that ${\boldsymbol{\phi}_{i}}$ is a $P$-dimensional phase delay vector in MOCSM rather than a scalar in MOSM. The phase delay vector provides a possibility that each dimension is allowed to has specific phase other than shared phase in MOSM. More detailed comparisons between SM-LMC, CSM, MOSM, and MOCSM in terms of hyper-parameters and degrees of freedom are given in Table \ref{tab:a}. In Table \ref{tab:a}, all LMC-based kernels use free form parameterization \cite{bonilla2008multi}, $\theta_{f}$ and $\vtime_{\ell}$ are the signal variance and length-scale in SE and Mat\'ern  kernel, respectively. Here we use $q$ instead of $i$ as a sub-index of a component. For SM-LMC, CSM, MOSM and MOCSM, $Q$ denotes the number of components, and $M$  the number of channels (outputs).

\begin{table*}[!ht]
	\caption{Comparisons between MOCSM and other kernels .} \label{tab:a}
\begin{center}
	\begin{tabular}{clll}
		\toprule    
		{Kernel} & {Parameters} & {Degrees of freedom} \\
		\midrule
		SE-LMC & $\{{B},\theta_{f},\vtime_{\ell}\}$ & $(M^2+M)/2+P+1$ \\
		Mat\'ern-LMC & $\{{B},\theta_{f}, \vtime_{\ell}\}$ & $(M^2+M)/2+P+1$\\
		SM-LMC & $\{{B}_{q},\,w_{q},\,\boldsymbol{\mu}_{q},\,\Sigma_{q}\,\}^{Q}_{q=1}$ & $Q((M^2+M)/2+2P+1)$ \\
		CSM & $\{\sigma^{q},\,\mu^{q},\,\{{w}^{q}_{r},\,{\phi}^{q}_{r},\,\phi^{1q}_{r}{\overset{\Delta}{=}}{0}\}^{M}_{r=1}\}^{Q}_{q=1}$ & $2Q+QM(2Q-1)$
		\\
		MOSM & $\{\{w_{m}^{q},\,\boldsymbol{\mu}_{m}^{q},\,\Sigma_{m}^{q},\,\boldsymbol{\theta}_{m}^{q},\,{\phi}_{m}^{q}\}_{m=1}^{M}\}_{q=1}^{Q}$ & $QM(3P+2)$ \\
		MOCSM & $\{\{w_{m}^{q},\,\boldsymbol{\mu}_{m}^{q},\,\Sigma_{m}^{q},\,\boldsymbol{\theta}_{m}^{q},\,\boldsymbol{\phi}_{m}^{q}\}_{m=1}^{M}\}_{q=1}^{Q}$ & $QM(4P+1)$  \\
		\bottomrule
	\end{tabular}
\end{center}
\end{table*}

\section{Experiments}\label{sec:exp}

In all experiments we consider the empirical spectral density ${\vs}$ as derived from the data, and then apply a Bayesian Gaussian mixture model (GMM) $p({{\Theta |\vs}})=\sum _{i=1}^{Q}{\tilde {w}_{i}}{\mathcal {N}}({\tilde{\boldsymbol{\mu}}_{i}},{\tilde {\Sigma} _{i}})$ in order to get the $Q$ cluster centers of the Gaussian spectral densities \cite{herlands2016scalable}.  We use the Expectation Maximization (EM)  algorithm \cite{Moon1997The} to estimate the parameters ${\tilde {w}_{i}}$, $\tilde{\boldsymbol{\mu }}_{i}$, and $ {\tilde {\Sigma} _{i}}$. The results are used as initial values of ${w}_{i}$, ${\boldsymbol {\mu}_{i}}$, and $ {\Sigma _{i}}$, respectively, for each channel in MOCSM. 

We used the same initialization and optimization method to infer the hyper-parameters of the considered kernels, the latter amounts to minimize the NLML  with respect to parameters chosen to construct the kernel, plus the noise hyper-parameters. Once the hyper-parameters are optimised, computing the predictive posterior involves the standard GP procedure with the joint covariances of the corresponding baseline kernel. 

We compare MOCSM with existing MOGP methods, namely SM-LMC, CSM, MOSM,  on an artificial dataset and on multiple real world datasets.  First, we show the ability of MOCSM in modeling 
correlated outputs simultaneously by considering a mixed signal sampled from a Gaussian distribution, its integral, and its derivative. 
Next, we use MOCSM for prediction tasks on real world problems with two benchmark datasets and three sensor array datasets \footnote{http://slb.nu/slbanalys/historiska-data-luft/} related to climate change and air pollution monitoring: temperature evolution, sliding mean ozone concentration, and global radiation extrapolation. For the artificial dataset and three sensor array datasets, we set $Q=10$ in MOCSM, MOSM, CSM, SM-LMC. Since these sensor datasets are positive signal, in theory GP could not be applied to this type of data. However in practice  the support of the GP for strictly positive data often will not have any significant mass for negative values. 
Therefore we consider this (and other) examples as practical real-life applications of GP's. 
As performance metric we consider the mean absolute error ${\mathrm {MAE} ={{\sum _{i=1}^{n}\left|y_{i}-\tilde{y}_{i}\right|}/{n}}}$. We implemented the model in Tensorflow \cite{Abadi2016c} and GPflow \cite{Matthews2017} because of its scalability and  automatic differentiation routines.  

\subsection{Artificial data: learning correlated mixed signals simultaneously}
We designed an artificial experiment inspired by \cite{Parra2017}, to model multiple non-linear correlated channels simultaneously. The three channels consist of a mixed signal, its integral, and its derivative. Here the numerical integral and derivative of a signal are respectively the area (trapezoidal rule, approximating the region under the graph of the function) and the slope of the tangent line to the graph of the function at two points, respectively. These are rough approximations of the exact integral and derivative when the interval of two points is not small enough. As a consequence, although the original signal is smooth, the numerical integral and derivative signals here considered do not retain smoothness.    The numerical integral and derivative of the signal is more complicated than the original generative signal, that's why in this setting channels are 
correlated and the derivative and integral signals contain more involved patterns. Specifically, the signal of the first channel is sampled from $\mathcal{GP}(0, K_\text{SM})$ ($Q=4$) with length 300 in the interval [-10, 10],  and then its first integral and derivative are numerically computed to be the signals of the second and third channel. This experiment is intended to validate the interpolation, extrapolation, and signal recovery ability of MOCSM as well as to compare its pattern recognition performance with that of other MOGP approaches. For the first channel, we randomly choose half of the data as training data, and the rest as test data. The integral signal in the interval [-10, 0] is used for training (in gray), the rest of the signal is used for testing (in green). The derivative of the signal in the interval [0, 10] is used for training and the rest of the signal is used for testing.

\begin{figure*}[!ht]
	\centering 	
\begin{subfigure}[t]{0.325\linewidth}
	\centering
	\includegraphics[width=1\linewidth]{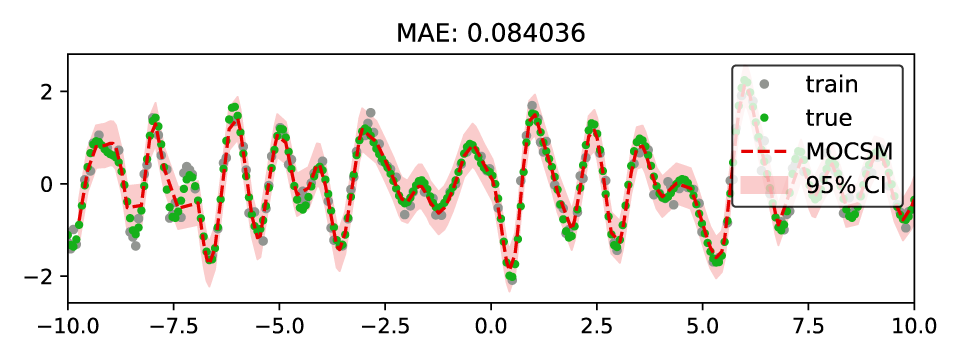}
\end{subfigure}
\begin{subfigure}[t]{0.325\linewidth}
	\centering
	\includegraphics[width=1\linewidth]{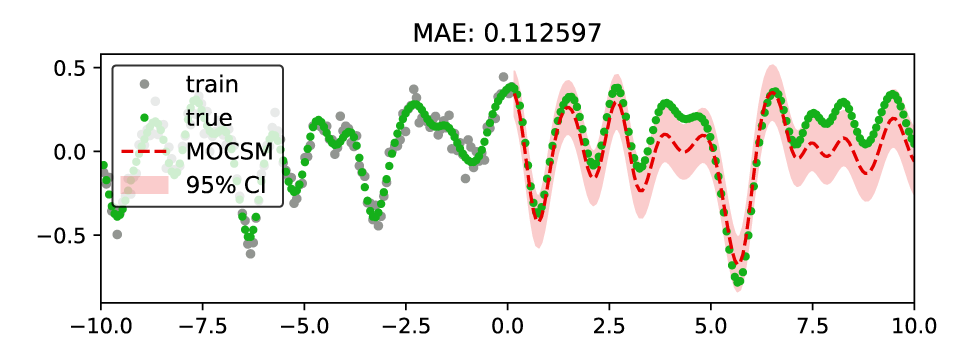} 
\end{subfigure}
\begin{subfigure}[t]{0.325\linewidth}
	\centering
	\includegraphics[width=1\linewidth]{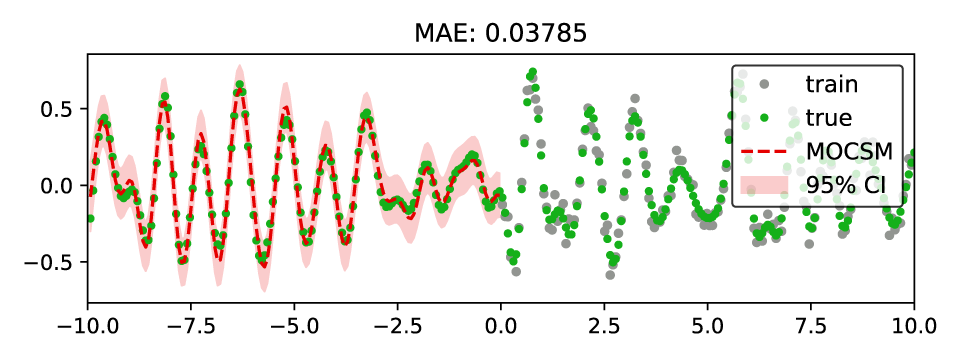} 
\end{subfigure}\\

\begin{subfigure}[t]{0.325\linewidth}
	\centering
	\includegraphics[width=1\linewidth]{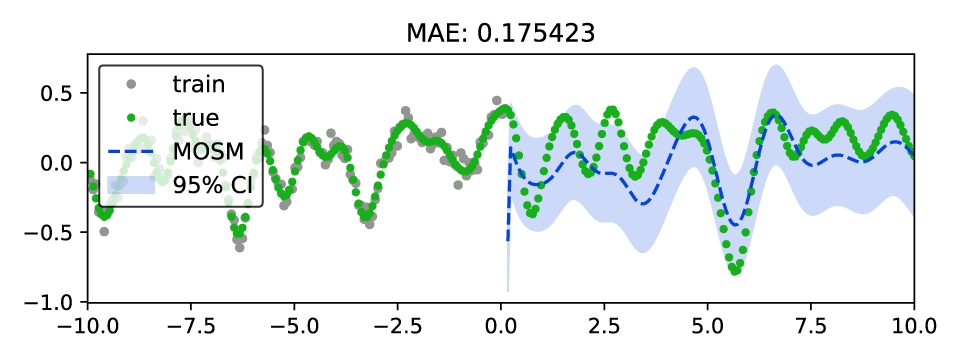} 
\end{subfigure}
\begin{subfigure}[t]{0.325\linewidth}
	\centering
	\includegraphics[width=1\linewidth]{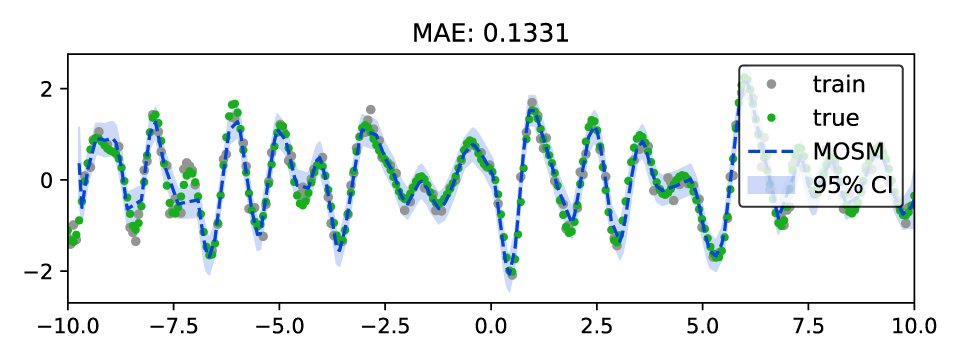} 
\end{subfigure}
\begin{subfigure}[t]{0.325\linewidth}
	\centering
	\includegraphics[width=1\linewidth]{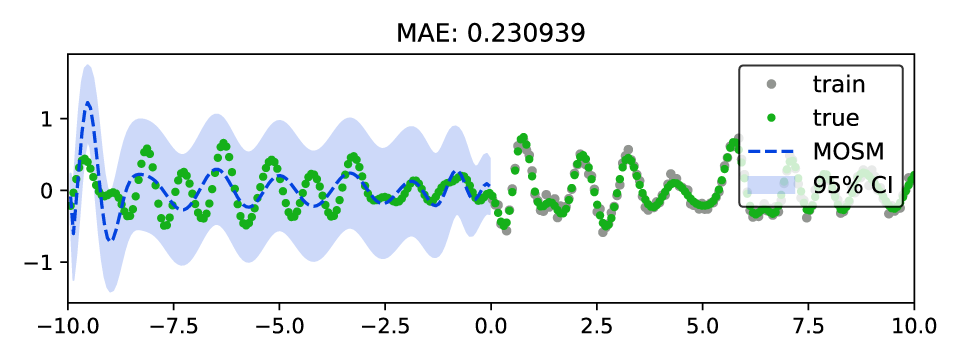} 
\end{subfigure}\\

\begin{subfigure}[t]{0.32\linewidth}
	\centering
	\caption{Signal $\sim\mathcal{GP}(0, K_\text{SM}(Q = 4))$} 
\end{subfigure}
\begin{subfigure}[t]{0.32\linewidth}
	\centering
	\caption{Integral of the signal}
\end{subfigure}
\begin{subfigure}[t]{0.32\linewidth}
	\centering
	\caption{Derivative of the signal} 
\end{subfigure}
\caption{Performance of MOCSM (in red dashed line) and MOSM (in blue dashed line) on artificial dataset. (a) Signal sampled from $\mathcal{GP}(0, K_\text{SM})$ with $Q=4$, training data are randomly chosen from the signal and the rest as test data. (b) Integral of the signal was numerically computed, the first half of data $x\in[-10, 0]$ was selected as a training and the remaining data as a test data. (c) Derivative of the signal was numerically computed, the last half of data $x\in[0, 10]$ was selected as a training and the rest as a test.}\label{fig:art1}
\end{figure*}

The performance of MOCSM (in dashed red line) on the first channel is shown in Figure \ref{fig:art1} (a). Although MOCSM performed slightly better, all mentioned GP methods learned the covariance and interpolated the missing values well. The extrapolation results on the integral and the derivative of the signal are shown in Figure \ref{fig:art1} (b) and Figure \ref{fig:art1} (c), respectively. The integral of the signal contains more  involved patterns than the original signal, which are difficult to identify and extrapolate. On these signals MOCSM performs better than the other baselines, with lowest MAE (see Table \ref{tab:t_exp1}) and smallest confidence interval. 

Predictions obtained using  SE-LMC and Mat\'ern-LMC kernels are of low quality, especially for the extrapolation tasks (integral and derivative signals): it is very hard for them to find valid patterns in the data, like the evolution of trend over time. That's a result of lack of periodic modeling ability in SE-LMC and Mat\'ern-LMC kernels, because they do not involve cosine and sine functions.

Overall, results on the artificial dataset indicate the capability of MOCSM to model integration and differentiation patterns of the generated signal simultaneously.

\subsection{Temperature extrapolation}

Sensor networks monitoring climate change and global warming in Stockholm provide real time surveillance, historical analysis, and future forecasting of the regional environment. Particularly extrapolating future temperature values  help to guide policy making and social development. We know that, as a result of global warming, the fluctuation of temperature becomes a sensitive topic with regard to balance of survival and development. On the other hand, temperature is one of main factors in affecting local air pollution levels because it determines chemical reaction and change of air pollutant. There are multiple global trends and local patterns which can shape temperature evolution. Global trends are natural evolution mechanisms within the global climate system itself. Local patterns are external forces caused by local surroundings and industry activities. Usually global trends affect long term evolution of temperature at a large scale. Local patterns always shape the short term and medium-term change of temperature at a small scale. Both global trends and local patterns are time and phase dependent and tightly coupled together. Here we use temperature recordings as a real world example to shows extrapolating ability of MOCSM in sensor networks. Based on the extrapolation results, suggestions may be given on measures to mitigate global warming.

\begin{figure*}[t]
	\centering 	
\begin{subfigure}[t]{0.325\linewidth}
	\centering
	\includegraphics[width=1\linewidth]{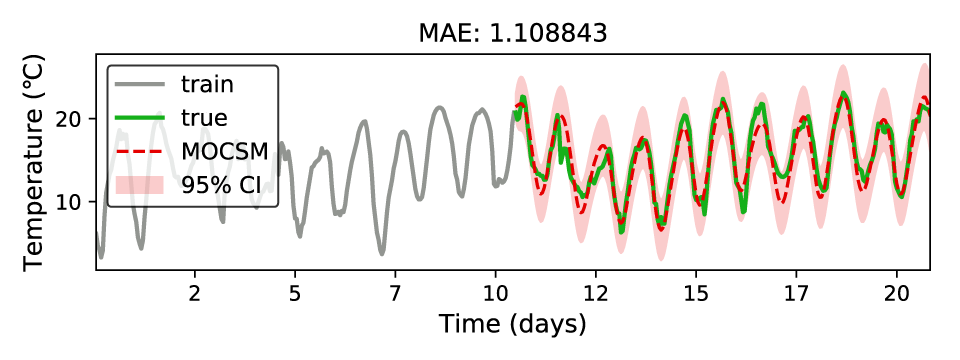} 
	\caption{MOCSM} 
\end{subfigure}
\begin{subfigure}[t]{0.325\linewidth}
	\centering
	\includegraphics[width=1\linewidth]{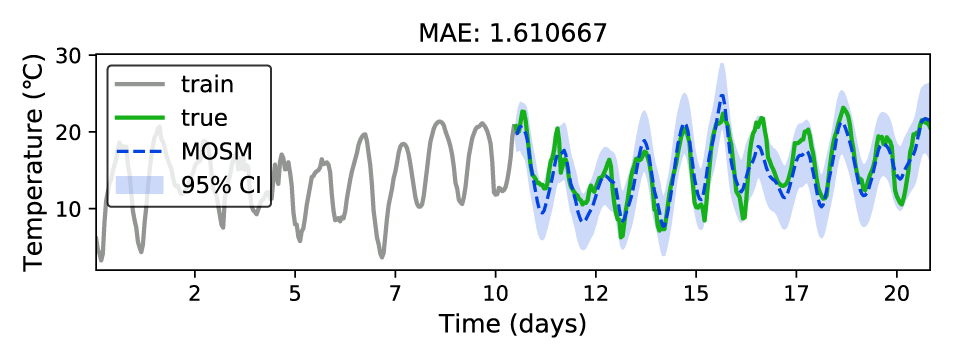} 
	\caption{MOSM} 
\end{subfigure}
\begin{subfigure}[t]{0.325\linewidth}
	\centering
	\includegraphics[width=1\linewidth]{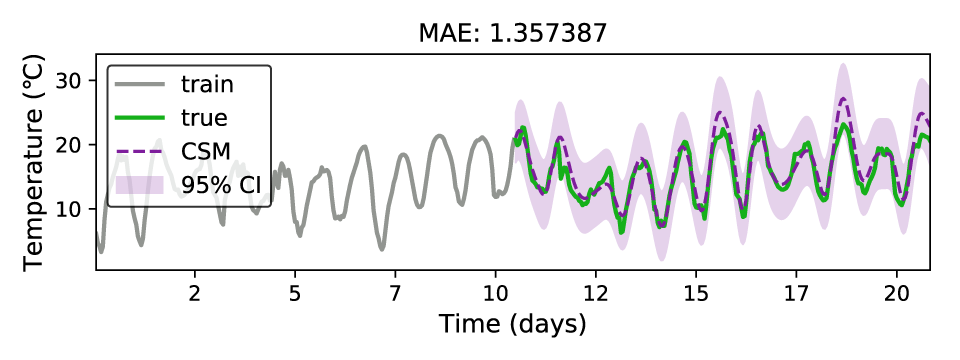} 
	\caption{CSM} 
\end{subfigure}\\
\begin{subfigure}[t]{0.32\linewidth}
	\centering
	\includegraphics[width=1\linewidth]{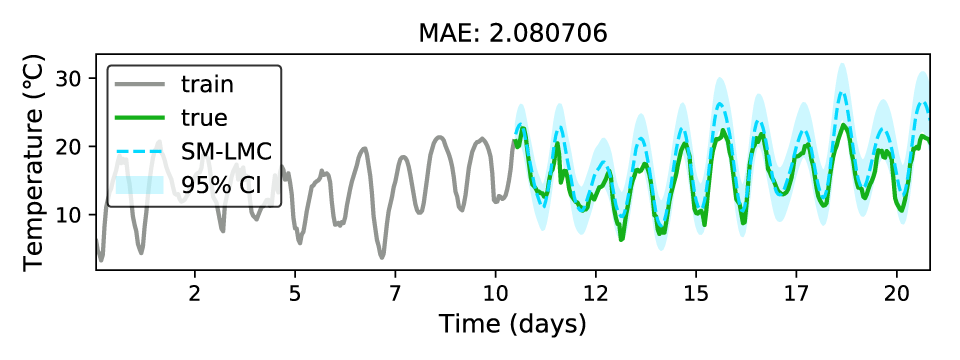} 
	\caption{SM-LMC} 
\end{subfigure}
\begin{subfigure}[t]{0.32\linewidth}
	\centering
	\includegraphics[width=1\linewidth]{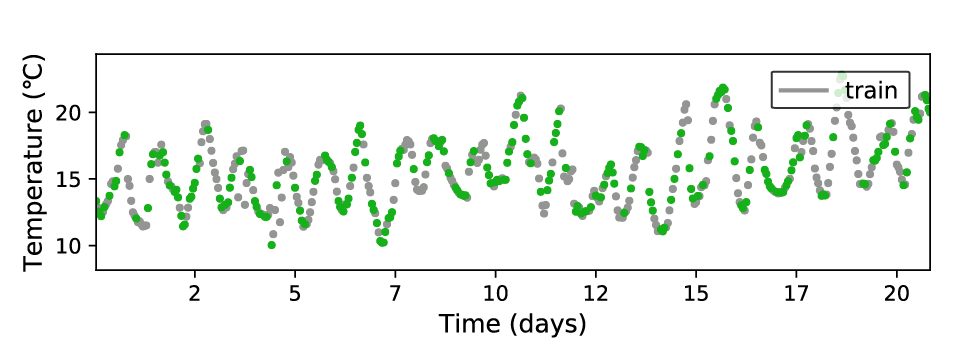} 
	\caption{Channel 1} 
\end{subfigure}
\begin{subfigure}[t]{0.32\linewidth}
	\centering
	\includegraphics[width=1\linewidth]{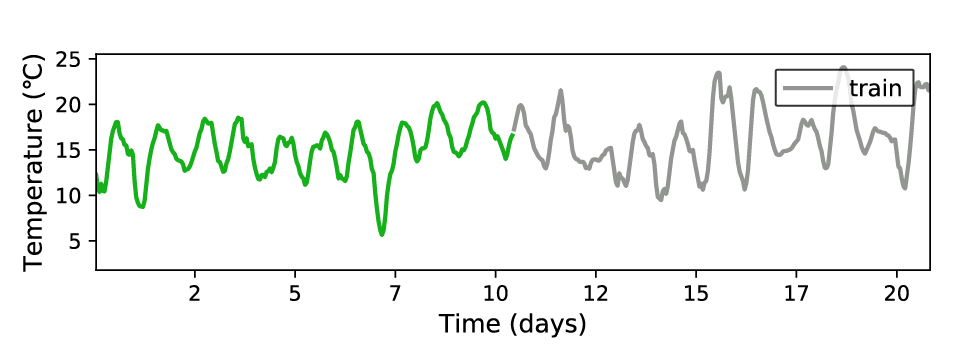} 
	\caption{Channel 3} 
\end{subfigure}
\caption{Temperature extrapolation. Performance comparison between MOCSM and recently proposed spectral mixture kernels: (a) MOCSM (in red dashed line), (b) MOSM (in blue dashed line), (c) CSM (in purple dashed line), (d) SM-LMC (in cyan dashed line), (e) training data of the 1th channel, (f) training data of the 3rd channel.}\label{fig:temp}
\end{figure*}

The temperature monitoring recordings are recorded from a number of stations (Torkel Knutssonsgatan, Marsta, Norr Malma) in Stockholm and outside. For instance: Torkel Knutssonsgatan's measurement at the urban background, Marsta's measurement at a high-altitude tower, Norr Malma's measurement at the regional background. We consider each station to be a channel: Torkel Knutssonsgatan as channel 1, Marsta as channel 2, and Norr Malma as channel 3. We observe from Figure \ref{fig:temp} that the change of temperature has an apparent oscillatory behavior. In this case, we use temperature time series from 22 June 2017 to 12 July 2017, in 1 hour intervals. Here we just focus on the task in channel 2. Specifically, a randomly chosen half of the temperature data in Torkel Knutssonsgatan, the first half of the temperature data in Marsta, and the last half of the temperature data in Norr Malma are used for training. The last half of temperature data in Marsta is used for testing. The change of temperature in each channel is affected by their 
interaction of time and phase related local patterns and global trends.

From Figure \ref{fig:temp}, as a result of time and phase dependent local patterns within channels (local patterns depend on surroundings) and time and phase dependent global trends between channels (global trends depend on seasonal or yearly factors), changes in temperature are fluctuating. MOCSM, however, consistently outperforms the other baselines and achieves lowest MAE and smallest predicted confidence interval (see Figure \ref{fig:temp} and Table \ref{tab:t_exp1}).

\subsection{Sliding mean Ozone concentration extrapolating}
Secondly, we use an 8-hour average Ozone concentrations dataset. The Ozone dataset comes from an air pollution monitoring sensor network in Stockholm. The Ozone concentration is recorded from 5 April, 2017 to 25 April, 2017, in one-hour intervals at Torkel Knutssonsgatan, Norr Malma, and Hornsgatan stations. As a result of moving average, noise was smoothed and therefore the evolution of ozone concentration is more smooth and less fluctuant. In other words, patterns from the frequency domain should become more clear. However, Ozone concentration depends more on local environment and human activities. Thus its extrapolations is more difficult than that of temperature. 

\begin{figure*}[!ht]
	\centering 	
\begin{subfigure}[t]{0.325\linewidth}
	\centering
	\includegraphics[width=1\linewidth]{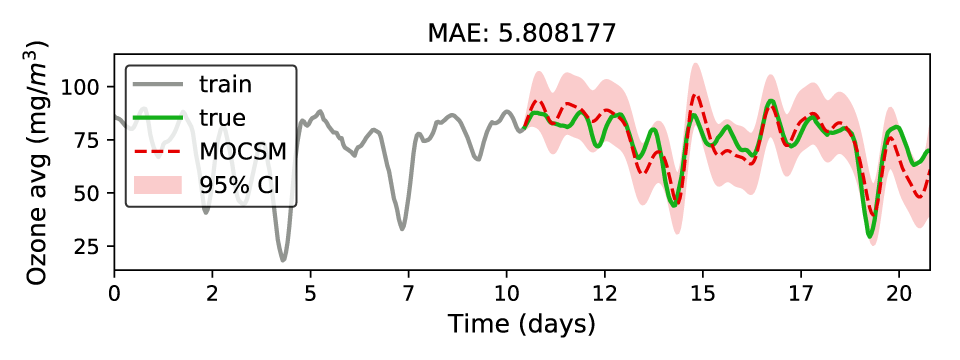} 
	\caption{MOCSM} 
\end{subfigure}
\begin{subfigure}[t]{0.325\linewidth}
	\centering
	\includegraphics[width=1\linewidth]{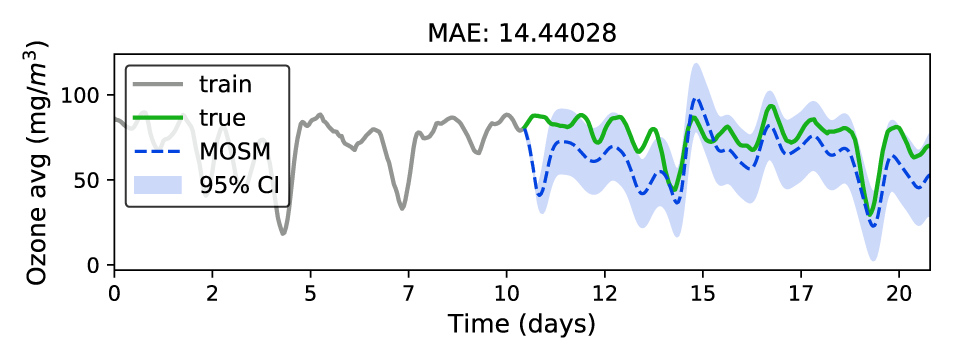} 
	\caption{MOSM} 
\end{subfigure}
\begin{subfigure}[t]{0.325\linewidth}
	\centering
	\includegraphics[width=1\linewidth]{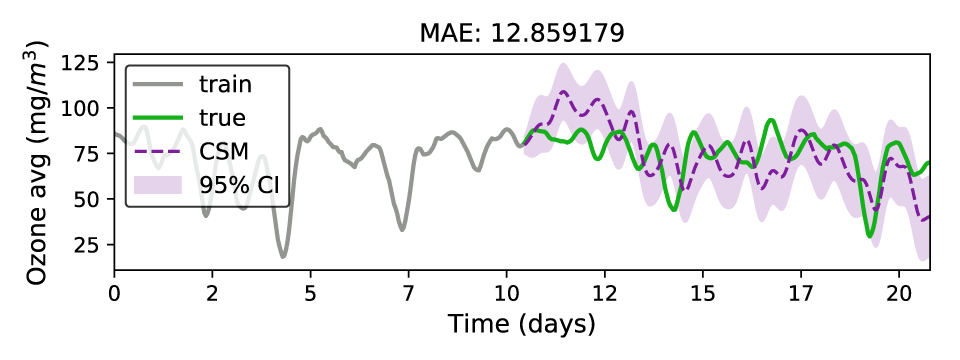} 
	\caption{CSM} 
\end{subfigure}\\
\begin{subfigure}[t]{0.325\linewidth}
	\centering
	\includegraphics[width=1\linewidth]{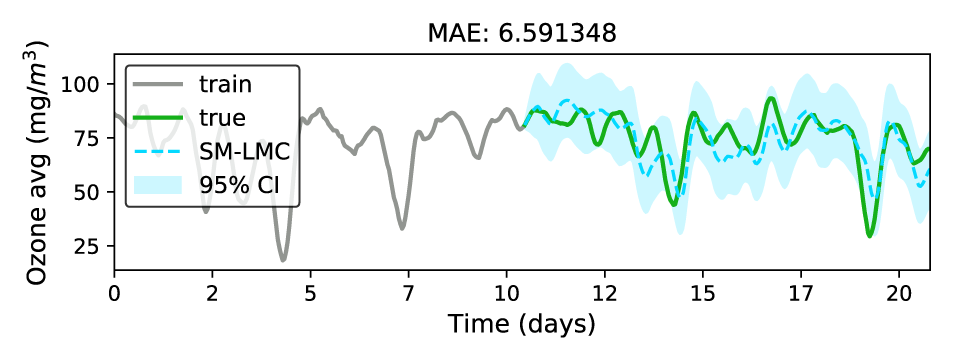} 
	\caption{SM-LMC} 
\end{subfigure}
\begin{subfigure}[t]{0.325\linewidth}
	\centering
	\includegraphics[width=1\linewidth]{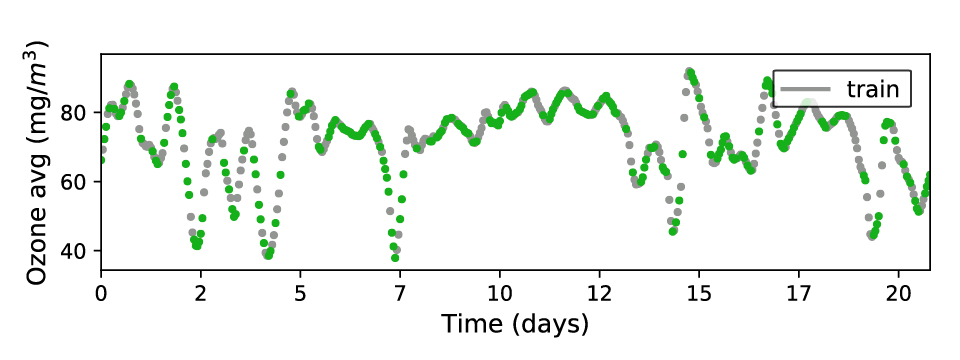}
	\caption{Channel 1} 
\end{subfigure}
\begin{subfigure}[t]{0.325\linewidth}
	\centering
	\includegraphics[width=1\linewidth]{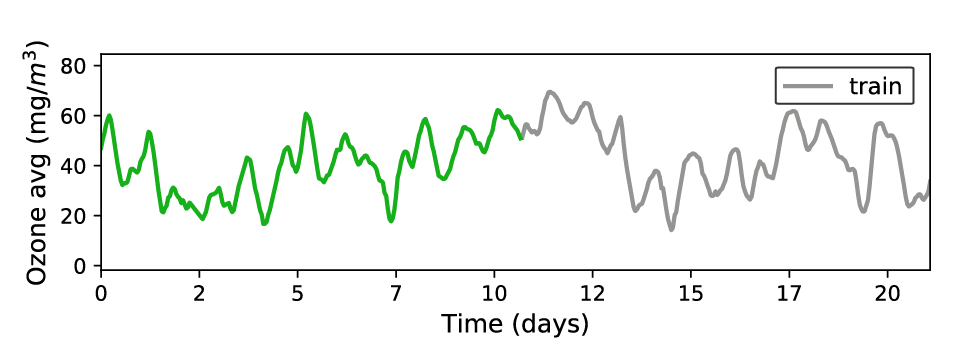}
	\caption{Channel 3} 
\end{subfigure}

\caption{Sliding mean Ozone concentration extrapolation. Performance comparison between MOCSM and recently proposed spectral mixture kernels: (a) MOCSM (in red dashed line), (b) MOSM (in blue dashed line), (c) CSM (in purple dashed line), (d) SM-LMC (in cyan dashed line), (e) training data of the 1th channel, (f) training data of the 3rd channel.}\label{fig:ozne}
\end{figure*}

Although the Ozone concentration data was smoothed by a sliding window of 8-hour values, local patterns in each channel depending on surroundings and global trends related to large-scale climate change still exist. Both local patterns and global trends are time and phase dependent over the period of recording. Modeling from multiple channels can benefit the 
extrapolating rather than learning one channel alone, because the evolution of Ozone concentration in each station is a result of 
interaction of these local patterns and global trends. Here  half of data was randomly chosen in Torkel Knutssonsgatan (channel 1), the first half of data in Norr Malma (channel 2), and the last half of data in Hornsgatan (channel 3) are used for training in channel 1, channel 2, and channel 3, respectively, and the remaining data in Norr Malma as testing data. In this case we still aim to extrapolate the long range trends of sliding mean Ozone concentration in Norr Malma. With the same setting as in the previous experiments, the performance of SM-LMC, CSM, MOSM, MOCSM are shown in Figure \ref{fig:ozne} and Table \ref{tab:t_exp1}.

As seen in Figure \ref{fig:ozne}, patterns in sliding mean Ozone concentration are more difficult and less clear to capture than temperature. It is even hard for human to find any quasi periodical or periodical trends. Although all MOGPs methods can extrapolate trends of sliding mean Ozone concentration, but MOCSM is achieves superior performance (see Figure \ref{fig:ozne} and Table \ref{tab:t_exp1}) and is able to correctly predict the appearance of all low peaks and all high peaks.  This experiment substantiates the strong pattern learning abilities MOCSM.

\subsection{Global radiation backward extrapolation}
After two extrapolating experiments, we conduct now a backward extrapolation on global radiation recordings of the sensor networks in order to validate the %long range 
back extrapolating capability and historical signal recovery ability of MOCSM and other baselines. The global radiation is an important parameter reflecting global climate evolution and change of atmosphere. Global radiation is the total amount of direct, diffuse, and reflected solar energy received by the Earth's surface, which is mainly affected by sun altitude and cloud cover. 

The change of sun altitude has a global influence on earth with a one-year period, while the cloud cover has a local influence with an a-periodic behavior. These global and local patterns show time and phase related variability over the period of recording. Usually for small scale variation (1 hour interval), the global radiation depends more on its surroundings and location because local weather condition determines the cloud cover. Empirical analysis shows various time and phase dependent characteristics of this global radiation: short term variations, medium term monthly patterns and non-strict periodic long term trends related to position of moon and sun, and some white noises. The time of appearance of high peak in global radiation is not periodical and its amplitude is always irregular.

\begin{figure*}[t]
	\centering 	
\begin{subfigure}[t]{0.325\linewidth}
	\centering
	\includegraphics[width=1\linewidth]{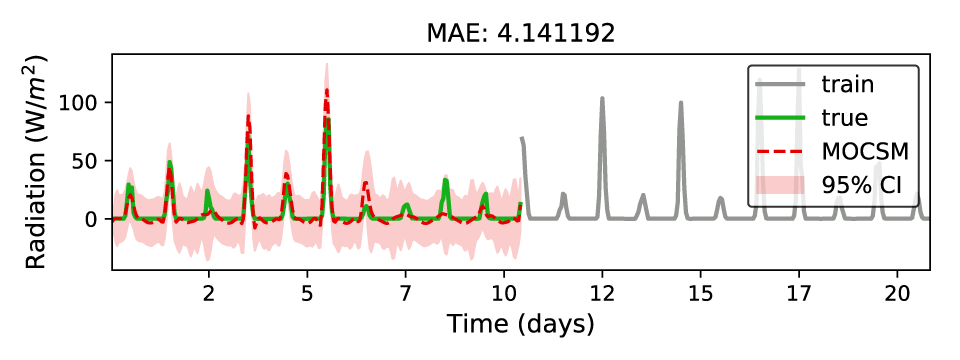} 
	\caption{MOCSM} 
\end{subfigure}
\begin{subfigure}[t]{0.325\linewidth}
	\centering
	\includegraphics[width=1\linewidth]{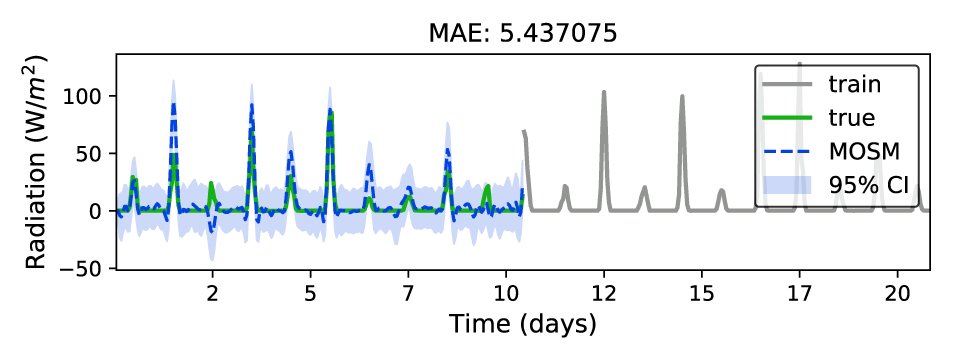} 
	\caption{MOSM} 
\end{subfigure}
\begin{subfigure}[t]{0.325\linewidth}
	\centering
	\includegraphics[width=1\linewidth]{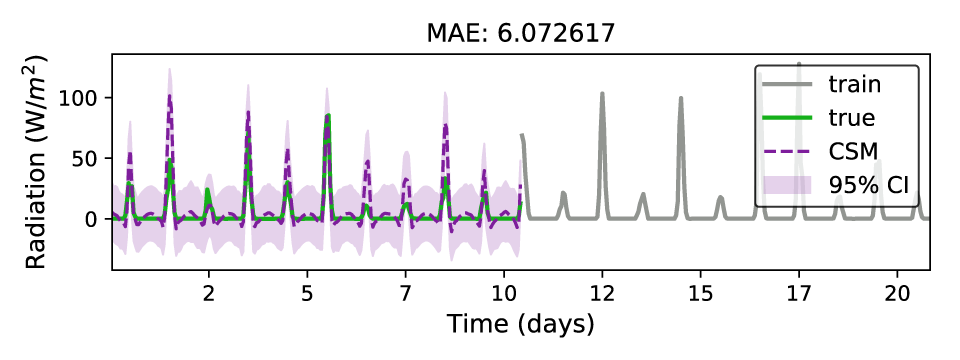} 
	\caption{CSM} 
\end{subfigure}\\
\begin{subfigure}[t]{0.325\linewidth}
	\centering
	\includegraphics[width=1\linewidth]{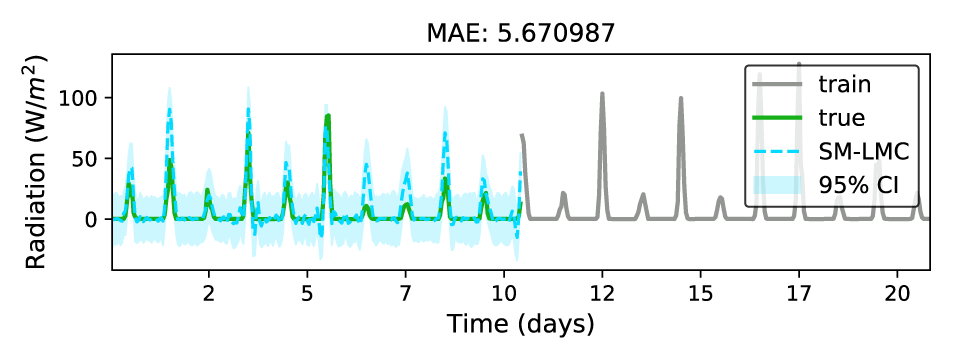} 
	\caption{SM-LMC} 
\end{subfigure}
\begin{subfigure}[t]{0.325\linewidth}
	\centering
	\includegraphics[width=1\linewidth]{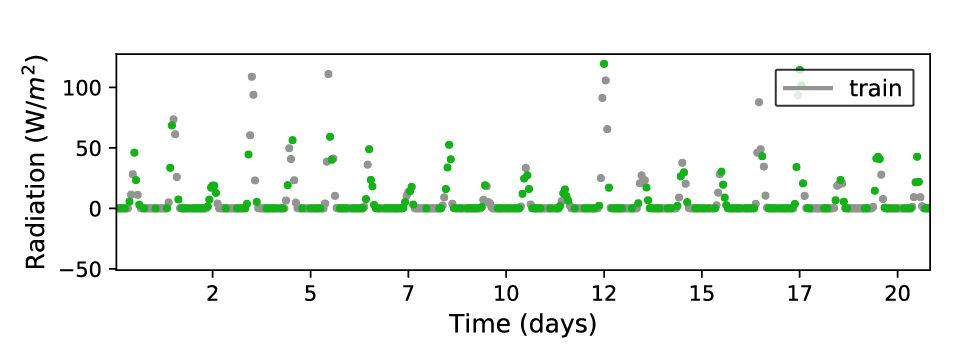} 
	\caption{Channel 1}
\end{subfigure}
\begin{subfigure}[t]{0.325\linewidth}
	\centering
	\includegraphics[width=1\linewidth]{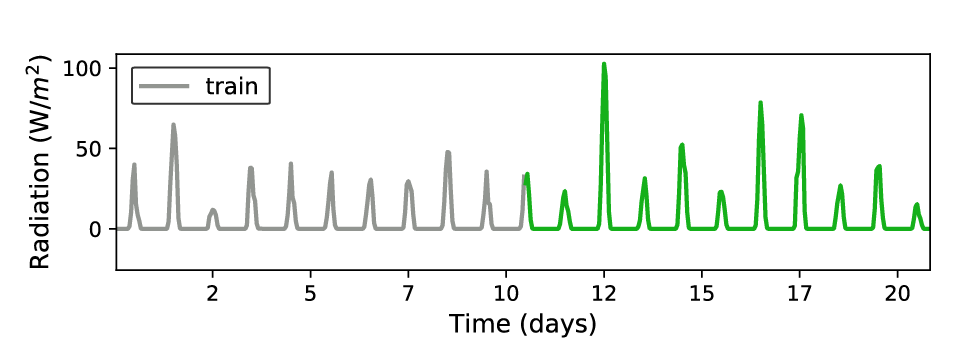} 
	\caption{Channel 2} 
\end{subfigure}
\caption{Global radiation back extrapolation. Performance comparison between MOCSM and recently proposed spectral mixture kernels: (a) MOCSM (in red dashed line), (b) MOSM (in blue dashed line), (c) CSM (in purple dashed line), (d) SM-LMC (in cyan dashed line), (e) training data of the 1th channel, (f) training data of the 2nd channel.}\label{fig:radiation}
\end{figure*}

The global radiation dataset is collected from three stations in Stockholm city: Torkel Knutssonsgatan's measurement at urban background, Marsta's measurement at a high-altitude tower, Norr Malma's measurement at regional background. All recordings cover 24 hours at 1 hour interval and missing values are filtered. In this case, we consider a global radiation recording from 5 December, 2017 to 26 December, 2017. Interestingly, the changing of global radiation over time looks like a non-continuous impulse signal because the sun radiation at night is almost equal to zero. From Figure \ref{fig:radiation} we can observe that the appearance of high peak in global radiation is irregular and its time of duration is short and fluctuant. Thus, complicated patterns contained in this non-continuous impulse signal are very difficult to be discovered using ordinary kernels. In such a multi-output scenario, the change of global radiation in each channel is caused by their 
interaction of time and phase related local and global patterns. 

In our experiment, we randomly chose half of global radiation data in Torkel Knutssonsgatan, the first half of global radiation data in Marsta, and the last half of global radiation data in Norr Malma as training data in channel 1, channel 2, and channel 3, respectively. Different from the first and second real world experiment, the rest of the  Norr Malma time series is here used for testing. In this setting, MOCSM consistently outperforms the other baselines with a lower MAE (see Figure \ref{fig:radiation} and Table \ref{tab:t_exp1}). Results indicate that all methods have difficulty to capture the trends of high peak appearance, and only MOCSM can forecast it without overestimation.

\begin{table*}[t]
	\caption{Performance of MOCSM and other kernels on artificial and real world datasets.}\label{tab:t_exp1}
\begin{center}
	\begin{tabular}{c r @{.} l r @{.} l r @{.} l r @{.} l r @{.} l r @{.} l}
		\toprule
		{Kernel} & \multicolumn{2}{c}{$\mathcal{GP}(0, K_\text{SM})$} & \multicolumn{2}{c}{Integral} & \multicolumn{2}{c}{Derivative} & \multicolumn{2}{c}{Temperature} & \multicolumn{2}{c}{Ozone} & \multicolumn{2}{c}{Radiation} \\
		\midrule
		SE-LMC   & 0&156      & 0&225     & 0&212   & 13&677 & 62&154 & 5&499\\
		Mat\'ern-LMC  & 0&126 & 0&219     & 0&221   & 14&737 & 70&867 & 5&230\\
		SM-LMC   & 0&124      & 0&195     & 0&195   &  2&081 &  6&591 & 5&671\\
		CSM      & 0&128      & 0&200     & 0&210   &  1&357 & 12&859 & 6&072\\
		MOSM     & 0&133      & 0&175     & 0&231   &  1&611 & 14&440 & 5&437\\
		MOCSM    & \textbf{0}&\textbf{084}  & \textbf{0}&\textbf{113} & \textbf{0}&\textbf{038} & \textbf{1}&\textbf{109} & \textbf{5}&\textbf{808} & \textbf{4}&\textbf{141} \\
		\bottomrule
	\end{tabular}
\end{center}
\end{table*}

Table \ref{tab:t_exp1} summarizes the performance of the considered kernel methods on the artificial  and the real world datasets. The MOCSM kernel consistently achieves the lowest MAE. Predictions obtained using the SE-LMC and Mat\'ern-LMC kernels are very bad especially for extrapolation (Integral, Derivative signals, Temperature, and Sliding mean Ozone concentration) even if the methods achieve a good MAE. With these kernels it is very hard  to find any valid pattern in the data. We use the temperature in Marsta, the sliding mean Ozone concentration in Norr Malma, and the radiation in Norr Malma as tasks to perform 
extrapolation.

\subsection{Comparison on benchmark datasets}
In this section we compare our baselines on the benchmark datasets used in MOSM \cite{Parra2017}, namely the UK climate \footnote{www.cambermet.co.uk} and the multivariate Jura datasets \cite{Parra2017,Goovaerts1997,Alvarez2009}. 

Experiments conducted with the same experimental setting as used in \cite{Parra2017} yield similar results as those reported in  \cite{Parra2017}, with very similar performance of the considered MOGP baselines.  
We observed that in the considered experimental setting it is not difficult to capture dependencies between channels because the training input data of the objective channel has an intersection with the training input data of the other three channels. 
Also, we observed that in that experimental setting the whole signal was normalized before splitting the signal into training and testing dataset. 

In order to obtain a more challenging setting, 
we use the original signal (so we do not normalize it) because the original signal recorded by the sensor has already been mixed with white noise. 
The UK climate dataset (from 12 March, 2017 to 16 March, 2017, in 5-minute intervals) collected records from a weather sensor network with four monitoring stations in the south England: Cambermet, Chimet, Sotonmet, and Bramblemet. Each station corresponds to a channel in our experiments.
The following sampling method \cite{Parra2017,Alvarez2009} is considered: the first half of signal in an objective sensor and respectively chosen $N=1000$ samples randomly from other three sensors are used for training, while the second half of signal in the objective sensor is used for testing.  
We consider such setup for each of the four channels, thus obtaining four experiments. In these experiments all baselines consider five spectral components. 

For the multivariate Jura dataset, we use the same experiment setting reported in \cite{Parra2017,Goovaerts1997,Alvarez2009}. The concentrations of Cadmium and Copper can be related to other variables which are easy to measure, such as Nickel and Zinc for Cadmium; and Lead, Nickel and Zinc for Copper. Experimental results on UK climate and Jura datasets are shown in Table \ref{tab:t_exp2}. Results in Table \ref{tab:t_exp2} indicate state of the art performance of MOCSM also on these benchmarks.  

\begin{figure*}[t]
	\centering 	
\begin{subfigure}[t]{0.245\linewidth}
	\centering
	\includegraphics[width=1\linewidth]{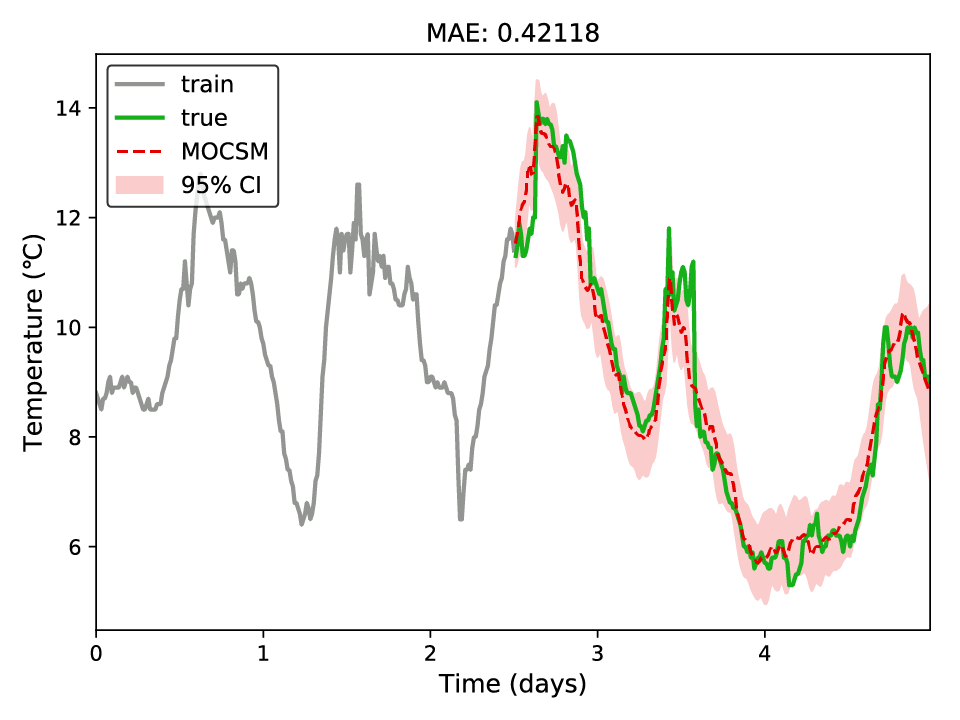} 
	\caption{MOCSM} 
\end{subfigure}
\begin{subfigure}[t]{0.245\linewidth}
	\centering
	\includegraphics[width=1\linewidth]{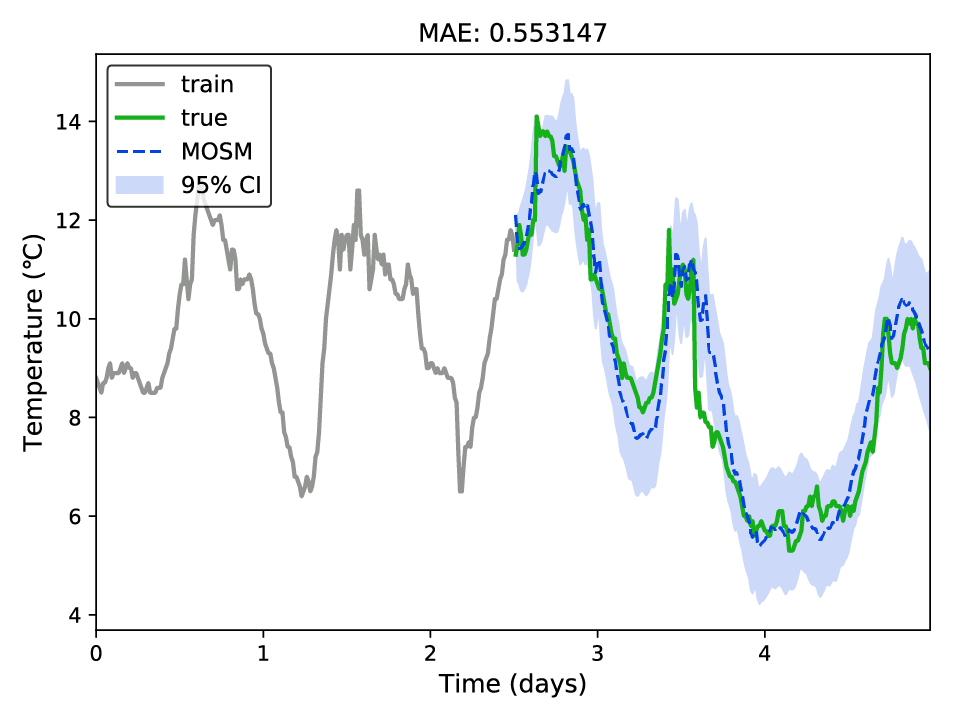} 
	\caption{MOSM} 
\end{subfigure}
\begin{subfigure}[t]{0.245\linewidth}
	\centering
	\includegraphics[width=1\linewidth]{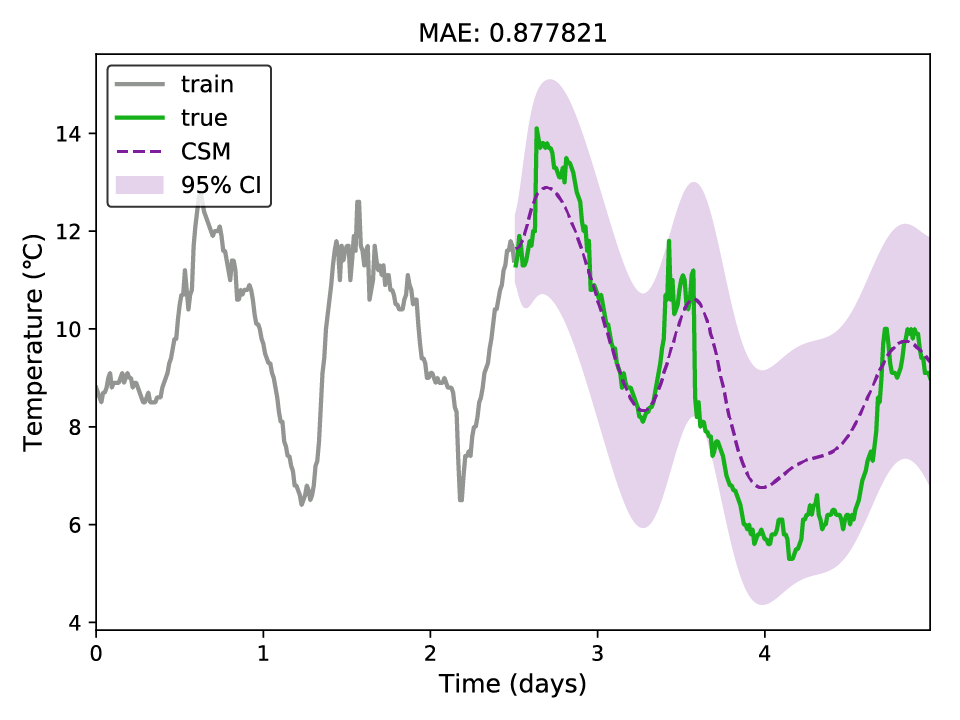} 
	\caption{CSM} 
\end{subfigure}
\begin{subfigure}[t]{0.245\linewidth}
	\centering
	\includegraphics[width=1\linewidth]{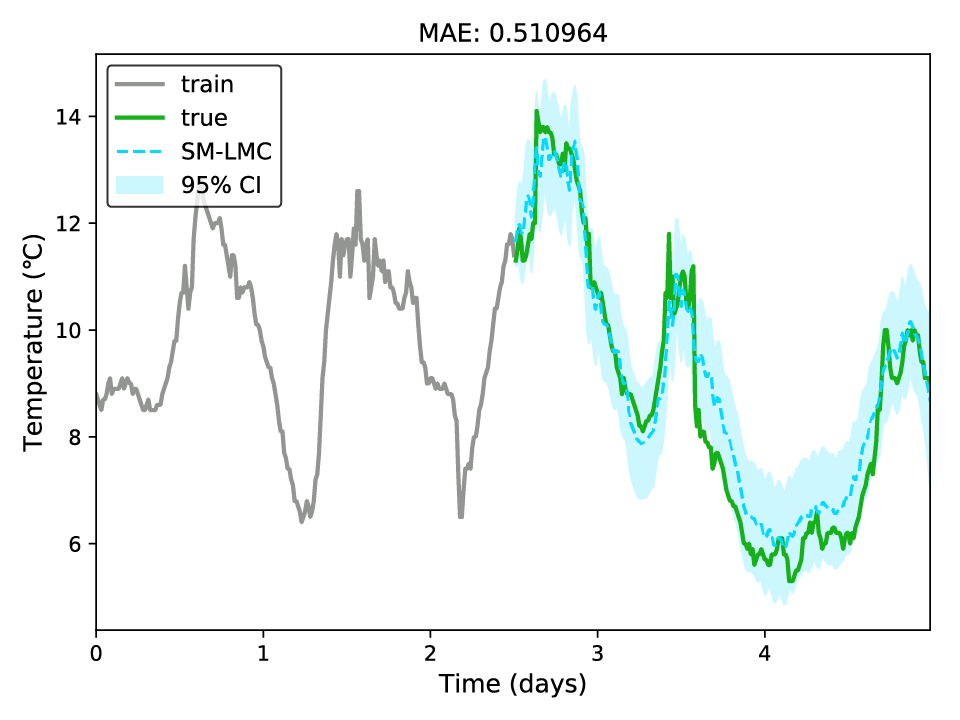} 
	\caption{SM-LMC} 
\end{subfigure}
\caption{UK climate signal missing input recovery. Performance comparison between MOCSM and other recently proposed spectral mixture kernels: (a) MOCSM (in red dashed line), (b) MOSM (in blue dashed line), (c) CSM (in purple dashed line), (d) SM-LMC (in cyan dashed line).}\label{fig:real4}
\end{figure*}

\begin{table*}[t]
	\caption{Performance of MOCSM and other kernels on UK climate and Jura datasets.}\label{tab:t_exp2}
\begin{center}
	\begin{tabular}{c r @{.} l r @{.} l r @{.} l r @{.} l r @{.} l r @{.} l}
		\toprule
		{Kernel} & \multicolumn{2}{c}{Cambermet} & \multicolumn{2}{c}{Chimet} & \multicolumn{2}{c}{Sotonmet} & \multicolumn{2}{c}{Bramblemet} & \multicolumn{2}{c}{Jura-Cadmium} & \multicolumn{2}{c}{Jura-Copper} \\
		\midrule
		SE-LMC   & 8&773     & 8&974     & 9&673   &  8&906 & 1&233 & 23&217\\
		Mat\'ern-LMC  & 7&677 & 7&983    & 8&704   & 8&073 & 1&234 & 16&276\\
		SM-LMC   & 0&510      & 0&506     & 1&067   &  0&709 &  0&460 & {7}&{000}\\
		CSM      & 0&877      & 0&851     & 1&658   &  0&562 & 0&470 & 7&400\\
		MOSM     & 0&553      & 0&572     & 0&988   &  0&587 & 0&430 & 7&300\\
		MOCSM    & \textbf{0}&\textbf{421}  & \textbf{0}&\textbf{487} & \textbf{0}&\textbf{835} & \textbf{0}&\textbf{549} & \textbf{0}&\textbf{418} & \textbf{6}&\textbf{823} \\
		\bottomrule
	\end{tabular}
\end{center}
\end{table*}

\section{Conclusion}\label{sec:con}
We proposed the generalized multi-output convolution spectral mixture kernel 
to describe cross covariance between multi-outputs in  a principled way. 
Results of extensive experiments indicated that MOCSM shows stronger abilities in modeling complicated 
correlation across multi-outputs than the considered baselines. Experiments on artificial datasets and real world datasets have shown that by using cross convolution of components within MOGPs, more irregular trends in the data can be recognized and learned and  long-term trends forecasting and 
back extrapolation can be performed in a more accurate way. 

MOCSM overcomes the problem that cross correlations are not so precise in MOSM, and extends the extrapolating ability of MOSM without increasing of computational complexity in the training and inference steps. In this work we did not address efficient inference of the MOCSM. At present, efficient inference approximation methods like FITC and PITC \cite{Alvarez2009, Alvarez2010,Alvarez2011, Williams2001, Quinonero-Candela2005, Snelson2006, Chalupka2013}, are not very effective for MOGPs because the labels of the inducing points are fixed and all outputs shared the same inducing points. 
Therefore interesting future research involves the development of sparse and efficient inference methods in a MOGP setting \cite{Rakitsch2013, Guarnizo2018, Liu2018a}.

Global optimization and initialization strategies of hyper-parameters are also very important for the MOGPs performance. Here we considered the empirical spectral density as derived from the data, and then applied a Bayesian Gaussian mixture model in order to initialize hyper-parameters.
More advanced initialization strategies, like \cite{Jang2017, Swersky2013,Martinez-Cantin2014,Knudde2017} need to be investigated in future work. 

\section*{Acknowledgment}
This work was financially supported by the National Key Research and Development Program of China (No.2017YFB0504203), Shenzhen Discipline Construction Project for Urban Computing and Data Intelligence, and China Scholarship Council. It was also partly financially supported by the Radboud University.

\ifCLASSOPTIONcaptionsoff
  \newpage
\fi



\bibliographystyle{IEEEtran}
\bibliography{main.bbl}

\end{document}